%% file: main.tex
\pdfoutput=1

\documentclass[11pt]{article}

\usepackage{acl}

\usepackage{times}
\usepackage{latexsym}

\usepackage{booktabs}
\usepackage{multirow}
\usepackage{xcolor}
\usepackage{colortbl}
\usepackage{array} 
\usepackage{amsmath}
\usepackage{xcolor}
\usepackage{makecell}
\usepackage{subcaption}

\newcolumntype{C}{>{\centering\arraybackslash}p{0.07\textwidth}}
\usepackage[T1]{fontenc}

\usepackage[utf8]{inputenc}

\usepackage{microtype}

\usepackage{inconsolata}

\usepackage{graphicx}

\title{DomainSum: A Hierarchical Benchmark for Fine-Grained Domain Shift in Abstractive Text Summarization}

\author{Haohan Yuan \qquad Haopeng Zhang\\
  University of Hawaii at Manoa \\
  \texttt{\{haohany,haopengz\}@hawaii.edu}}

\begin{document}
\maketitle
\begin{abstract}
Most research on abstractive summarization focuses on single-domain applications, often neglecting how domain shifts between documents affect performance and the generalization ability of summarization models. To address this issue, we introduce DomainSum, a hierarchical benchmark designed to capture fine-grained domain shifts in abstractive summarization. We categorize these shifts into three levels: genre, style, and topic, and demonstrate through comprehensive benchmark analysis that they follow a hierarchical structure. Furthermore, we evaluate the domain generalization capabilities of commonly used pre-trained language models (PLMs) and large language models (LLMs) in in-domain and cross-domain settings. Our benchmark and source code will be released at~\url{https://github.com/hpzhang94/DomainSum}.

\end{abstract}

\input{content/1introduction}

\input{content/2related}
\input{content/3method}
\input{content/4experiment}


\input{content/5analysis}
\section{Conclusion}
This paper presents DomainSum, a hierarchical benchmark for fine-grained domain shifts in abstractive summarization. We categorize domain shifts into genre, style, and topic levels and verify that genre shifts cause more variation, while topic shifts lead to more uniform changes through comprehensive multidimensional benchmark analysis. We also evaluate PLMs and LLMs on in-domain and cross-domain settings, showcasing their domain generalization abilities. DomainSum provides a valuable testbed for future research on summarization model generalization.

\section*{Limitations}

For the construction of our large-scale benchmark, we primarily reused existing datasets to represent various levels of domain shifts in DomainSum, rather than collecting new data, which would have been costly. The datasets we included are high-quality and publicly available for summarization tasks.

In our fine-tuning experiments, we sampled 10,000 training instances, 500 validation instances, and 500 test instances from the respective training, validation, and test sets of each dataset, rather than using the entire dataset. This decision was made to ensure fairness across datasets, as they vary in size. Previous research efforts~\cite{goyal2022news,zhang2024benchmarking} have also tested GPT-3 on similarly small subsets.

Lastly, we did not include domain adaptation methods~\cite{wang2019exploring,fabbri2020improving,laskar2022domain} in our experiments. The main goal of this paper is to provide a benchmark for evaluating fine-grained domain shifts in summarization. Exploring techniques for adapting summarization models to new domains is left for future work, with our benchmark serving as a valuable resource for this purpose.

\bibliography{custom}

\appendix
\input{content/6appendix}

\end{document}

%% file: content/1introduction.tex
\section{Introduction}
Abstractive summarization is a crucial task in natural language processing (NLP) that aims to generate concise and coherent summaries by interpreting and distilling essential information from a source text. As the volume of publicly available text continues to grow exponentially, the demand for automatic summarization methods has become more urgent. Recent advancements in pre-trained language models (PLMs) and large language models (LLMs) have significantly improved the performance of abstractive summarization systems, achieving unprecedented results in generating human-like summaries~\cite{liu2019text,liu2022brio,zhang2020pegasus,zhang2023summit}.

However, much of the current research focuses on summarizing specific types of documents, such as news articles or academic papers~\cite{zhang2023extractive, zhang2024systematic}. This narrow focus limits the models' ability to generalize across documents with diverse characteristics, hindering their effectiveness in real-world applications. Specifically, many models struggle to adapt to different content types due to unaddressed distributional discrepancies in the training data, a phenomenon commonly referred to as summarization domain shift~\cite{wang2019exploring}.

Recently, researchers have begun investigating how domain-specific corpus characteristics affect summarization performance. Early studies~\cite{hua2017pilot} examined out-of-domain training for summarization models or utilized document categories and latent topics for multi-task learning~\cite{cao2017improving}. \citet{wang2019exploring} modeled summarization domain shift by analyzing distributional differences between news sources (e.g., CNN vs. New York Times), applying meta-learning and domain tagging to tackle multi-domain extractive summarization. Building on this, \citet{yu2021adaptsum} explored domain adaptation for abstractive summarization, emphasizing the distributional disparities across document types (e.g., emails vs. news vs. dialogues) and proposed continued pre-training for low-resource settings. More recently, \citet{afzal2024adapteval} evaluated large language models (LLMs) for domain adaptation in zero-shot and few-shot settings, focusing on distributional differences across topics (e.g., science vs. medicine vs. government).


\input{figures/Domain_Shift_Documents}

Despite these efforts, a clear and consistent definition of domain shift in summarization remains elusive, and current methods lack a fine-grained measurement to capture the variability of domain shifts across different content types. As illustrated in Fig~\ref{fig:domain_shift}, the domain shift between two news articles on different topics should be smaller than the shift between a news article and a Reddit post. However, existing studies and benchmarks fail to capture this nuanced distinction in domain shifts across varying types of content.

To address these limitations, we introduce DomainSum, a hierarchical benchmark designed to investigate fine-grained domain shifts in abstractive summarization. Inspired by the systematic formulation of language style~\cite{dimarco1993computational}, we categorize domain shift into three distinct levels of granularity: genre shift, style shift, and topic shift, as depicted in Fig~\ref{fig:hierarchies}. We validate this categorization through a comprehensive corpus analysis, examining key properties such as compression ratio, density, and abstractiveness across these three levels. DomainSum is constructed from high-quality public datasets and encompasses five distinct domains for each level, creating a large-scale, hierarchical testbed. Furthermore, we evaluate the performance of both current PLMs and LLMs on our benchmark, uncovering their ability to handle varying degrees of domain shift. The key contributions of this paper are threefold:

\begin{itemize}
    \item We categorize summarization domain shift into three levels (genre, style, and topic) and present DomainSum, a large-scale hierarchical benchmark that enables a comprehensive evaluation of summarization performance across diverse content types.
    \item We perform a detailed analysis of domain shift within DomainSum across different granular levels, examining eight domain characteristic measures such as compression, density, coverage, diversity, and abstractiveness across domains.
    \item We evaluate the domain shift capabilities of existing PLMs and LLMs, highlighting their performance variations across different granular levels of domain shift.
\end{itemize}

%% file: figures/Domain_Shift_Documents.tex
\begin{figure}[t!]
    \centering
    \includegraphics[width=0.98\linewidth]{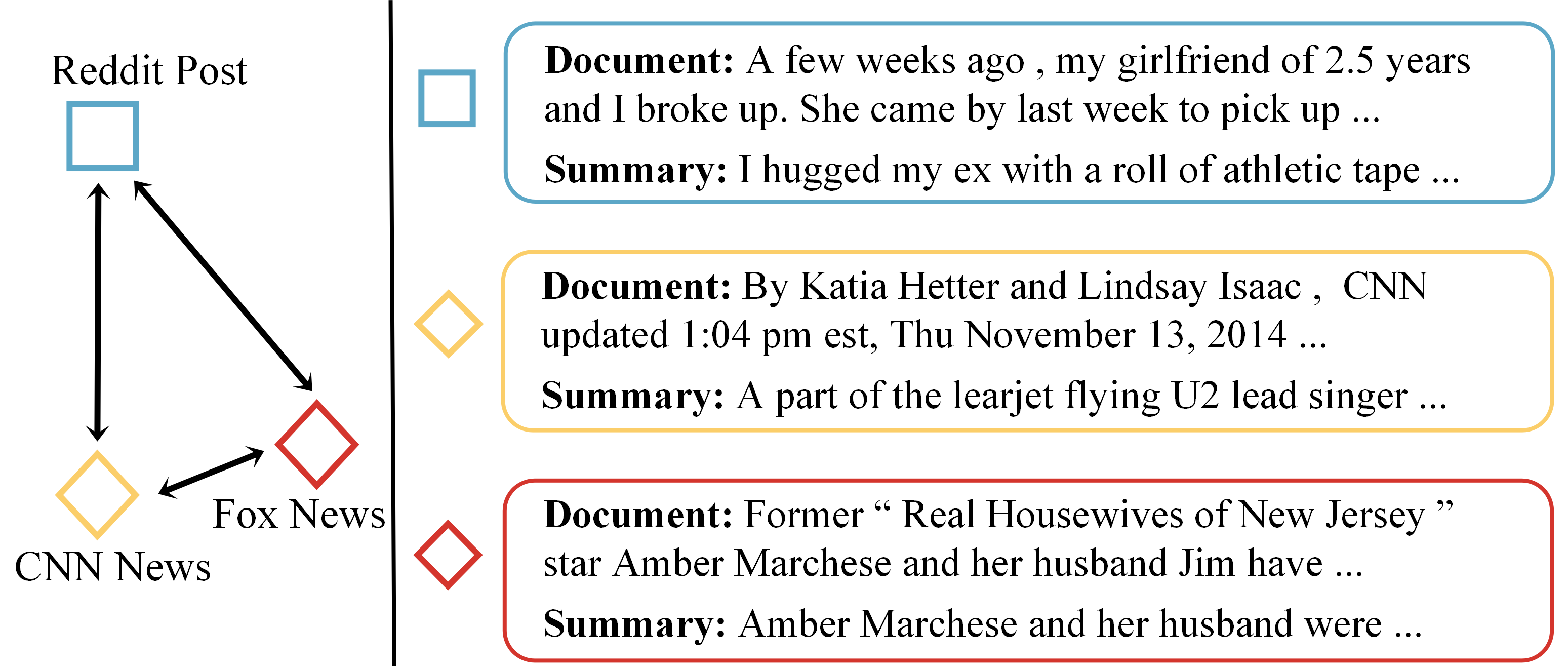}
    \caption{An illustration of varying domain shifts between content styles, with smaller shifts between news articles (CNN vs. Fox) and larger shifts between news articles and Reddit posts.
    }
    \label{fig:domain_shift}
\end{figure}

%% file: content/2related.tex
\section{Related Work}
\input{figures/Fig2_hierarchies}

\subsection{Abstractive Summarization}
Abstractive summarization aims to create concise and coherent summaries from scratch, often employing sequence-to-sequence models~\cite{sutskever2014sequence}. 
In contrast to extractive summarization methods~\cite{zhang2022hegel, zhang2023contrastive, zhang2023diffusum}, abstractive summarization offers greater flexibility and fluency while reducing redundancy. The advent of PLMs has significantly advanced the field, leading to notable improvements in fluency, coherence, and informativeness~\cite{lewis2019bart,zhang2020pegasus,liu2022brio}.

Recently, LLMs have gained considerable attention for summarization due to their strong in-context learning (ICL) capabilities~\cite{brown2020language} and chain-of-thought reasoning abilities~\cite{wei2022chain,wang2023element}. With minimal or no examples, LLMs can generate summaries that rival those produced by fine-tuned PLMs~\cite{min2022rethinking,afzal2024adapteval}. For example, \citet{goyal2022news} found that while GPT-3 summaries yielded slightly lower ROUGE scores, human evaluators preferred them. Similarly, \citet{zhang2024benchmarking} reported that LLM-generated summaries in the news domain performed comparably to those written by humans. Furthermore, \citet{zhang2023summit} introduced an iterative framework for text summarization, allowing LLMs to refine their outputs through self-evaluation and feedback, thereby improving both faithfulness and control.

\subsection{Domain Shift for Summarization}
Domain shift has been extensively studied in NLP and computer vision (CV)~\cite{ganin2016domain, gururangan2020don, ramponi2020neural}, focusing on the distributional differences between training and test data. However, exploration of domain adaptation in abstractive summarization remains limited. \citet{hua2017pilot} first investigated the adaptation of neural summarization models to out-of-domain data. Following this, \citet{wang2019exploring} examined domain shifts in news sources specifically for extractive summarization. \citet{magooda2020abstractive} expanded this area by introducing cross-domain data synthesis methods for summarization. \citet{yu2021adaptsum} addressed the adaptation of abstractive summarization across domains through continued pre-training. More recently, \citet{afzal2024adapteval} explored the application of LLMs for zero-shot and few-shot adaptation across various topics. Additionally, \citet{li2024word} found that the learning difficulty of datasets exhibits an almost linear relationship with cross-domain overlap and performance gain.

Despite these advancements, there is still a lack of clear and consistent definitions of domain shift in summarization, as well as a significant gap in studying the nuanced influences of document distributional differences on summarization.


%% file: figures/Fig2_hierarchies.tex
\begin{figure*}[t!]
    \centering
    \includegraphics[width=0.8\textwidth]{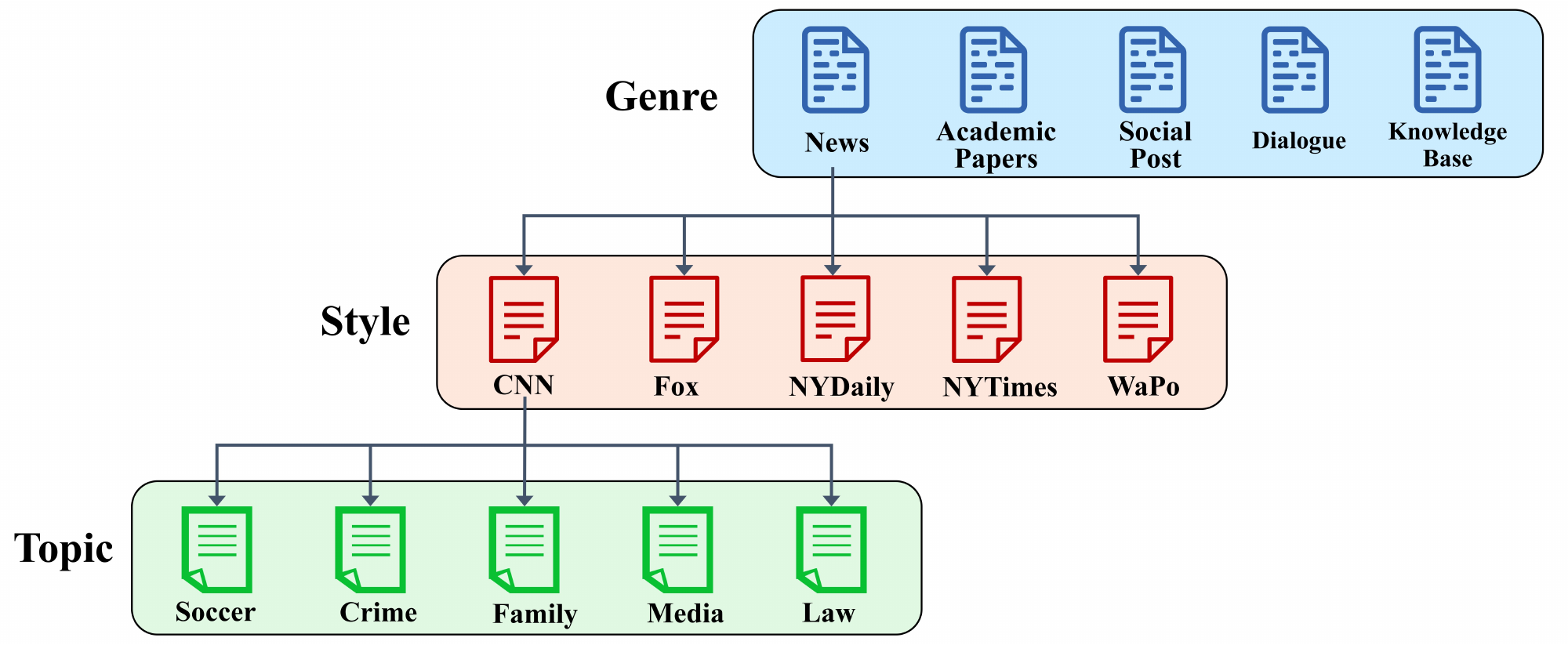}
    \caption{The overall hierarchical structure of DomainSum, featuring three granular levels of shifts (genre, style, and topic) and five distinct domains at each level.}
    \label{fig:hierarchies}
\end{figure*}

%% file: content/3method.tex
\section{DomainSum Benchmark}
This section introduces our DomainSum benchmark in detail. In Section~\ref{sec:construction}, we outline the benchmark's construction process, which leverages high-quality public datasets. Additionally, we present a comprehensive analysis of DomainSum using eight summarization domain characteristic measures in Section~\ref{sec:statistics}.

\subsection{Benchmark Construction}
\label{sec:construction}

Domain refers to a metadata attribute that divides data into distinct parts with varying distributions~\cite{joshi2012multi}. Domain shift is crucial for supervised systems, as a mismatch between training and testing domains can adversely affect performance. In the context of abstractive summarization, a conditional generation task, domain shift captures the distributional differences between document-summary pairs. Specifically, we propose to categorize domain shifts in summarization into three granular levels: \textbf{genre shift, style shift, and topic shift}, drawing inspiration from the systematic formulation of language style~\cite{dimarco1993computational}.

To capture the nuanced distinctions in domain shifts across different types of content, we construct our DomainSum benchmark hierarchically. Specifically, we create document-summary pairs from five distinct domains at each granular level, resulting in a $3\times5$ benchmark structure, as illustrated in Fig~\ref{fig:hierarchies}. The details of the data included are as follows:

\paragraph{Genre Shift} We begin by defining genre shift, which refers to the changes in content, format, and structure that arise when summarizing texts across different genres, such as news articles, academic papers, and fiction. This phenomenon presents distinct challenges in the summarization process, requiring models to adapt to the diverse conventions and audience expectations inherent to each genre.

For DomainSum, we leverage high-quality, publicly available datasets from five specific genres:
\textbf{i) News:} We pick the widely-adopted CNN/DailyMail~\cite{hermann2015teaching} news summarization dataset, which contains news articles alongside human-written highlights as summaries.
\textbf{ii) Academic papers:} We use the Arxiv dataset~\cite{cohan2018discourse} as a representative for scientific papers and long-form documents. Following the setup in~\cite{zhong2020extractive}, the introduction section is treated as the document and the abstract as the summary.
\textbf{iii) Social post:} We include the Reddit dataset~\cite{kim2018abstractive}, which contains highly abstractive summaries from social media. We specifically use the TIFU-long version, where the body text of a post is regarded as the document and the TL;DR as the summary.
\textbf{iv) Dialogue:} The SAMSum dataset~\cite{gliwa2019samsum} is incorporated to represent text message-like dialogues. It comprises single-document conversations created by linguists fluent in English, covering a broad range of formality levels and topics.
\textbf{v) Knowledge base:} We include the WikiHow dataset~\cite{koupaee2018wikihow}, which consists of diverse instructional contents extracted from an online knowledge base. Each summary succinctly encapsulates the main steps or advice provided in the guide article.

\input{tables/dataset}

\paragraph{Style Shift} As one finer-grained step, even within the same genre, different sources or authors produce articles with varying styles, distinguished by differences in tone, length, word frequency, and polarization. We define this disparity between documents and their summaries as style shift.

To capture this variation, we focus on the news article genre and construct fine-grained domain shift data that reflects diverse styles of news articles. Following the settings in~\cite{wang2019exploring}, we reuse and sample from the Newsroom corpus~\cite{grusky2018newsroom}, which comprises article-summary pairs authored by writers and editors from $38$ major publications between 1998 and 2017. We selected the top five publications (The New York Times (\textbf{NYTimes}), The Washington Post (\textbf{WaPo}), Fox News (\textbf{Fox}), the New York Daily News (\textbf{NYDaily}), and CNN (\textbf{CNN})) and processed the data in standardized formats. We specifically retained CNN news from the Newsroom dataset to ensure consistency in summaries across both genre and topic levels.

\paragraph{Topic Shift}
As a further step, documents within the same genre and style can still exhibit significant variation in terms of topics and emphasis. For instance, news articles may cover subjects such as sports or finance, incorporating different terminologies and catering to distinct audience expectations and summary information density. 

To account for this variation, we closely examine the news genre and the aforementioned CNN news style, constructing a finer-grained domain shift that captures diverse topics within the news article-summary pairs. Specifically, we employ the Latent Dirichlet Allocation (LDA) topic model~\cite{blei2003latent} to cluster the CNN news data by topic. From this analysis, we identify and include the five most frequent categories: \textbf{Soccer, Crime, Family, Media, and Law}. Additionally, we conduct human verification to manually filter out data with low topic relevance.

Altogether, the data instances from the three levels of summarization in DomainSum are organized in a hierarchical structure, capturing varying levels of granularity. Detailed statistics for the DomainSum benchmark, including the training, validation, and testing sets, are presented in Table~\ref{tab:dataset}.

\input{figures/Radar_Comparison}

\subsection{Benchmark Analysis} 
\label{sec:statistics}
Following the three granular levels defined in DomainSum, we conduct a comprehensive analysis of the benchmark using eight key measures to examine the data distribution and summarization style shift characteristics at each level. 

\subsubsection{Summarization Domain Characteristic Measures}
Given a set of document-summary paris $(D_t,S_t)$ from a specific domain t, we employ the following detialed summarization domain characteristic measures. For n-grams, we select $n=3$, which covers unigrams, bigrams, and trigrams in all measures.
\paragraph{Length} This measure includes both the average token length of documents $|D_t|$ and the average token length of summaries $|S_t|$.

\paragraph{Compression} Compression measures the ratio of document length to summary length, indicating how much information has been condensed through summarization~\cite{grusky2018newsroom}. \begin{align} \mathrm{Compression}(D_t,S_t)= \frac{|D_t|}{|S_t|}. \end{align}

\paragraph{Density} Density measures the degree to which a summary is directly extracted from its corresponding document~\cite{grusky2018newsroom}.  \begin{align} \mathrm{Density}(D_t, S_t) = \frac{1}{|S_t|} \sum_{f \in \mathcal{F}(D_t, S_t)} |f|^2, \end{align}
where $\mathcal{F}(D_t,S_t)$ is the set of extractive fragments shared between the document and the summary. It is formally defined as:

\paragraph{Diversity}
Diversity measures the variety of word usage within both the document and the summary~\cite{yogatama2015extractive}. It is calculated separately for the document and the summary as the average proportion of unique n-grams. Document Diversity is calculated as:
\begin{align}
    \mathrm{Document\ Diversity}(D_t) = \frac{1}{n} \sum_{i=1}^{n} \frac{|U_i(D_t)|}{|G_i(D_t)|},
\end{align}
where  $U_i(D_t)$  represents the set of unique n-grams in the document  $D_t$ , and  $G_i(D_t)$  is the total set of n-grams in the document. Similarly, Summary Diversity is calculated as:
\begin{align}
    \mathrm{Summary\ Diversity}(S_t) = \frac{1}{n} \sum_{i=1}^{n} \frac{|U_i(S_t)|}{|G_i(S_t)|}.
\end{align}

\paragraph{Coverage}
Coverage measures the proportion of n-grams from the document that appear in the summary~\cite{yogatama2015extractive}.
\begin{align}
\mathrm{Coverage}(D_t,S_t) = \frac{1}{n} \sum_{i=1}^{n} \frac{|G_i(S_t) \cap G_i(D_t)|}{|G_i(S_t)|}.
\end{align}

\paragraph{Abstractiveness}
Abstractiveness measures the extent to which a summary diverges from direct extraction by employing novel words or phrases~\cite{narayan2018don}. We calculate abstractiveness based on novel n-grams as:
\begin{align}
    \mathrm{Abstractiveness}(D_t, S_t) = \frac{1}{n} \sum_{i=1}^{n} \frac{|N_i(S_t, D_t)|}{|G_i(S_t)|},
\end{align}
where $N_i(S_t, D_t)$  represents the set of novel n-grams in the summary that do not appear in the document  $D_t$ .

\input{figures/VocabularyOverlap}

\subsubsection{Data Distribution Analysis}
Here we conduct analysis on DomainSum based on the above measures, and the overall results are presented in Figure~\ref{fig:radar}.

\paragraph{Genre Level} As shown in Fig~\ref{fig:genre_shift}, CNN/DM exhibits the highest density and coverage, aligning with its news feature where summaries frequently retain substantial information from the original text. PubMed produces the longest summaries and achieves the highest abstractiveness, likely due to the complexity and specificity inherent in biomedical research articles. In contrast, Reddit demonstrates the highest compression ratio, reflecting the informal and concise nature of user-generated social post. SAMSum, which primarily comprises dialogues, demonstrates the greatest diversity in summaries, likely resulting from the varied and informal language structures typical of conversational data.

\paragraph{Style Level} As illustrated in Fig~\ref{fig:style_shift}, news articles from different publishers exhibit markedly different styles. The Washington Post (WaPo) demonstrates the highest document length, compression, and abstractiveness, indicating that its summaries condense a substantial amount of content while introducing significant new material beyond the original text. Fox showcases the greatest summary diversity, reflecting a higher degree of linguistic variation in its summaries. In contrast, NYDaily ranks highest in coverage and density, as its summaries retain the most information directly from the original documents.

\paragraph{Topic Level}  We further examine news articles from CNN covering different topics, as illustrated in Fig~\ref{fig:topic_shift}. We observe that the Family category has the highest document length and compression, indicating that its summaries condense a substantial amount of content. The Crime category demonstrates the highest density, signifying that its summaries are closely aligned with the original text. The Law category shows the highest coverage, capturing more aspects of the original news in its summaries. Conversely, Soccer exhibits the highest abstractiveness, suggesting that its summaries introduce a greater amount of new information beyond the original text. Both the Media and Soccer categories display higher summary diversity, indicating a larger variety of language in their summaries.

\paragraph{Cross Level Comparison}

We also analyze the domain shift differences across levels. In terms of density, the Genre Shift shows high variability while the Style Shift displays more consistent density across categories, and the Topic Shift has a relatively uniform density distribution. Similarly for coverage, the Genre Shift demonstrates significant variation among categories, while the Style Shift presents a more evenly distributed coverage.

Overall, the Topic Shift shows the most consistent and balanced distribution across measures, while the Genre Shift displays the most extreme variations in density and coverage. The Style Shift falls in between, exhibiting moderate variations. These findings suggest that \textbf{topic shifts result in more uniform changes across text characteristics, whereas genre shifts lead to more dramatic variations, with style shifts producing intermediate effects}. This aligns with the \textbf{hierarchical design} of DomainSum and supports our hypothesis to categorize summarization domain shifts into hierarchical genre, style, and topic shifts.

\input{tables/Main-results}

\subsection{Vocabulary Overlap}
We also examine vocabulary overlaps between different domains across the three levels in DomainSum. As shown in Fig~\ref{fig:VOMAP}, the style shift shows the highest overlap between different news styles, consistently exceeding 75\%, likely due to shared source characteristics. In contrast, the genre shift level has lower overlap percentages, mostly falling below 60\%, indicating more distinct vocabulary usage across genres. The topic shift level reveals mixed overlap across news topics.

%% file: tables/dataset.tex
\begin{table}[t]
\centering
\small  
\begin{tabular}{l|ccc}
\hline
\textbf{Datasets} & \textbf{Train} & \textbf{Validation} & \textbf{Test} \\
\hline

\multicolumn{4}{c}{\textbf{\textsl{Genre Shift}}} \\
\hline
CNN/DM   & 287,084  & 13,367 & 11,489  \\
PubMed   & 83,233   & 4,946  & 5,025   \\
Reddit   & 41,675   & 645    & 645     \\
SAMSum   & 14,732   & 818    & 819     \\
WikiHow  & 168,126  & 6,000  & 6,000   \\
\hline
\multicolumn{4}{c}{\textbf{\textsl{Style Shift}}} \\
\hline
CNN &43,466 &4,563 &4,619  \\
Fox &78,760 &8,423 &8,392  \\
NYDaily &55,653 &6,057 &5,904  \\
NYTimes  &152,959 &16,488 &16,620  \\
WaPo  &95,379 &9,939 &10,072  \\
\hline
\multicolumn{4}{c}{\textbf{\textsl{Topic Shift}}} \\
\hline
Soccer &16,313 &1,889 &1,546  \\
Crime &24,597 &1,131 &973  \\
Family &26,019 &1,176 &1,047  \\
Media &24,364 &1,189 &919  \\
Law &15,564 &550 &532  \\
\hline
\end{tabular}

\caption{Overview of the train, validation, and test splits for DomainSum.
}
\label{tab:dataset}
\end{table}

%% file: figures/Radar_Comparison.tex
\begin{figure*}[ht]
    \centering
    \begin{minipage}[c]{0.32\linewidth}
        \centering
        \includegraphics[width=\textwidth]{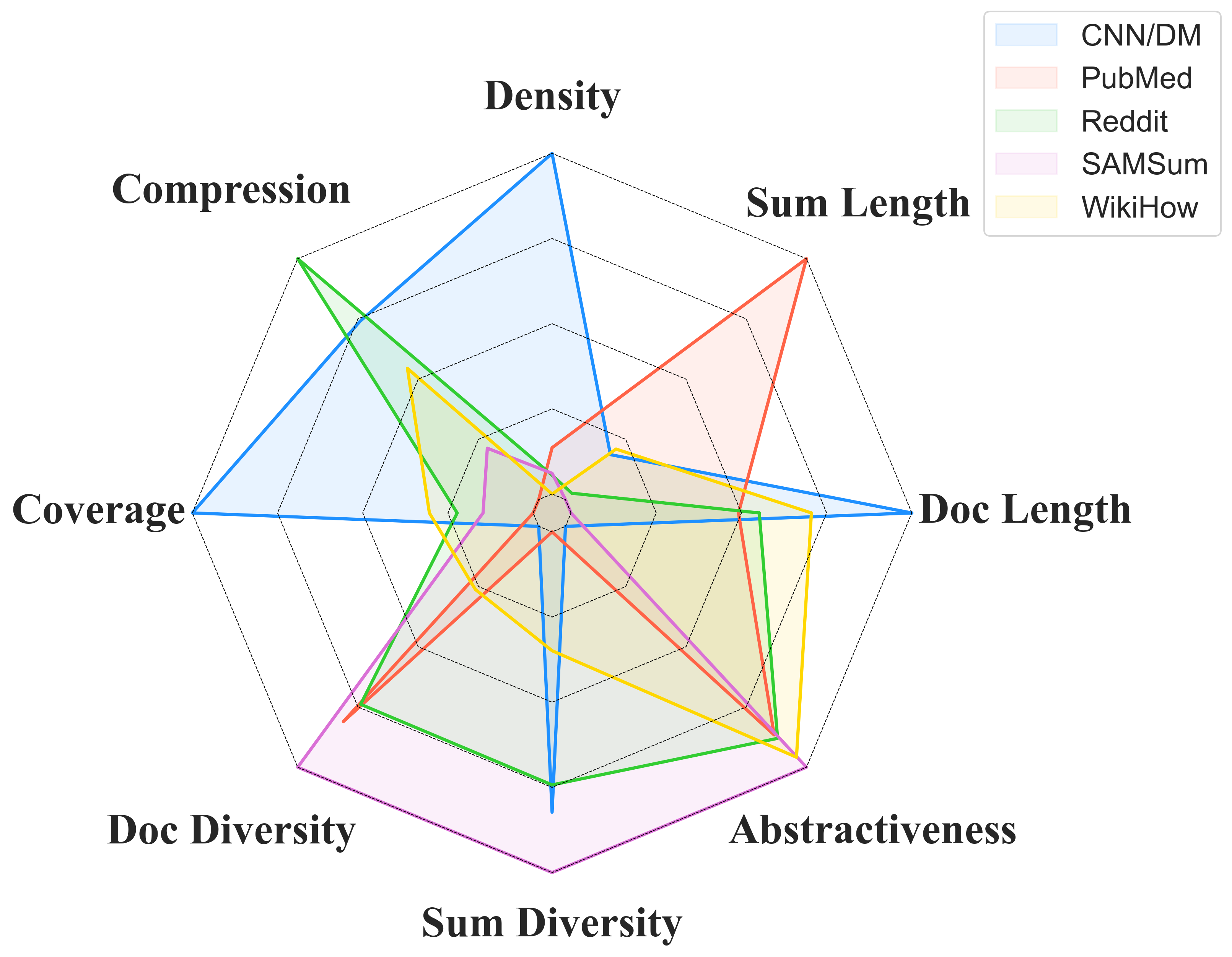}
        \subcaption{Genre Shift}
        \label{fig:genre_shift}  
    \end{minipage}
    \begin{minipage}[c]{0.32\linewidth}
        \centering
        \includegraphics[width=\textwidth]{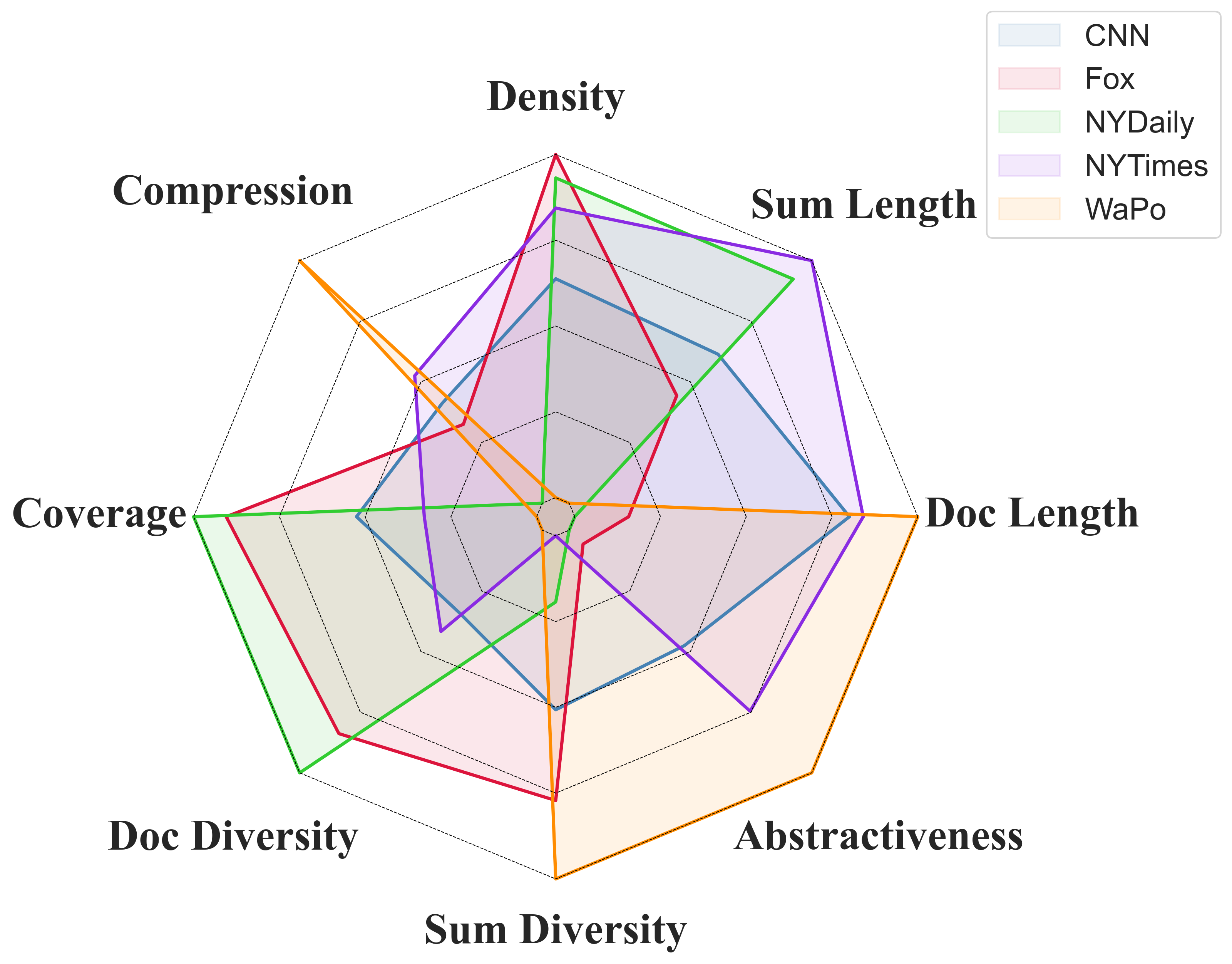}
        \subcaption{Style Shift}
        \label{fig:style_shift}  
    \end{minipage}
    \begin{minipage}[c]{0.32\linewidth}
        \centering
        \includegraphics[width=\textwidth]{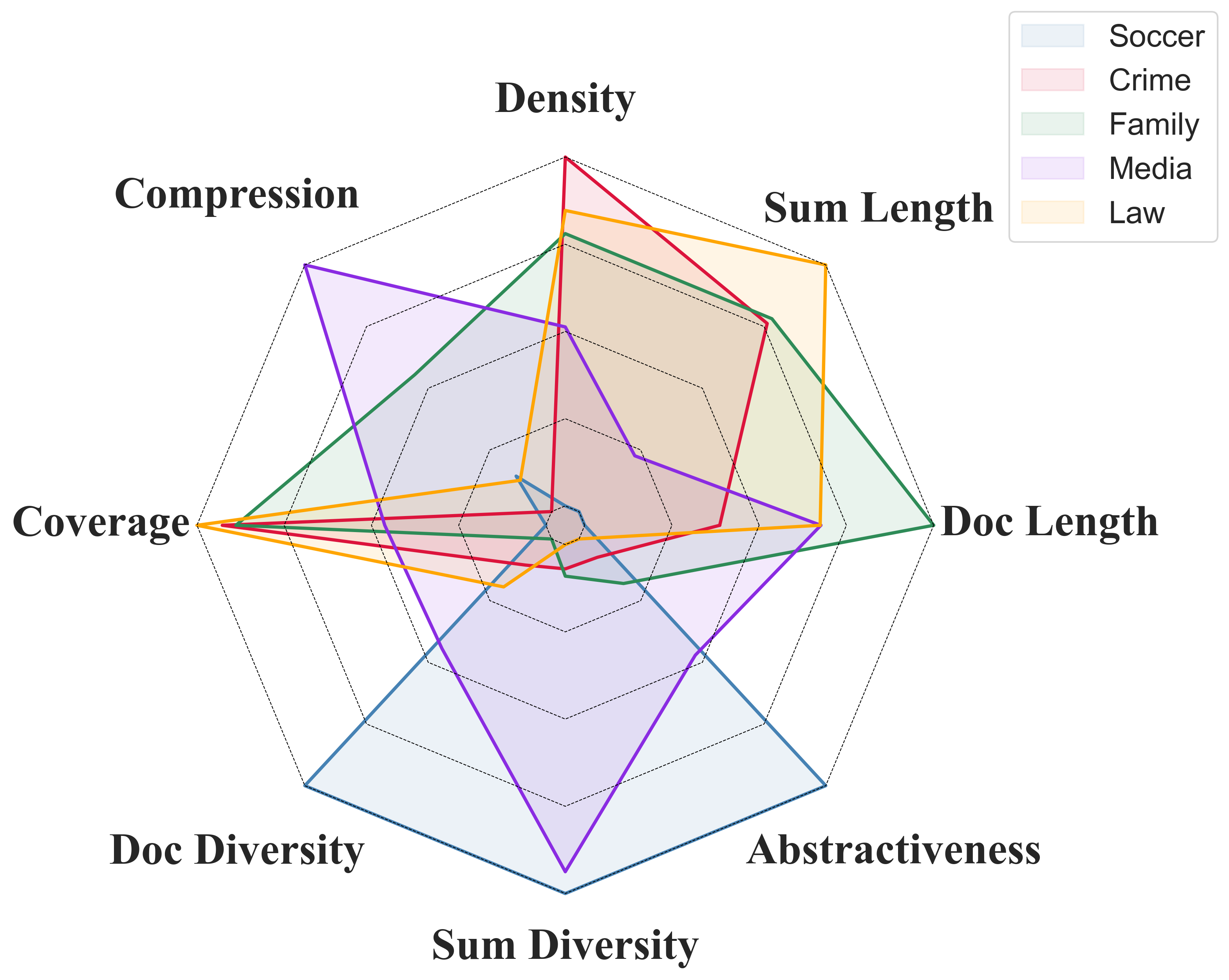}
        \subcaption{Topic Shift}
        \label{fig:topic_shift}  
    \end{minipage}
    \caption{Radar charts illustrating multi-dimensional benchmark analysis based on domain characteristic measures across three levels. The measures evaluated include document/summary length, density, compression, coverage, abstractiveness, and document/summary diversity.}
    \label{fig:radar}
\end{figure*}

%% file: figures/VocabularyOverlap.tex
\begin{figure*}[ht]
	\centering
	\begin{minipage}[c]{0.32\linewidth}
		\centering
		\includegraphics[width=\textwidth]{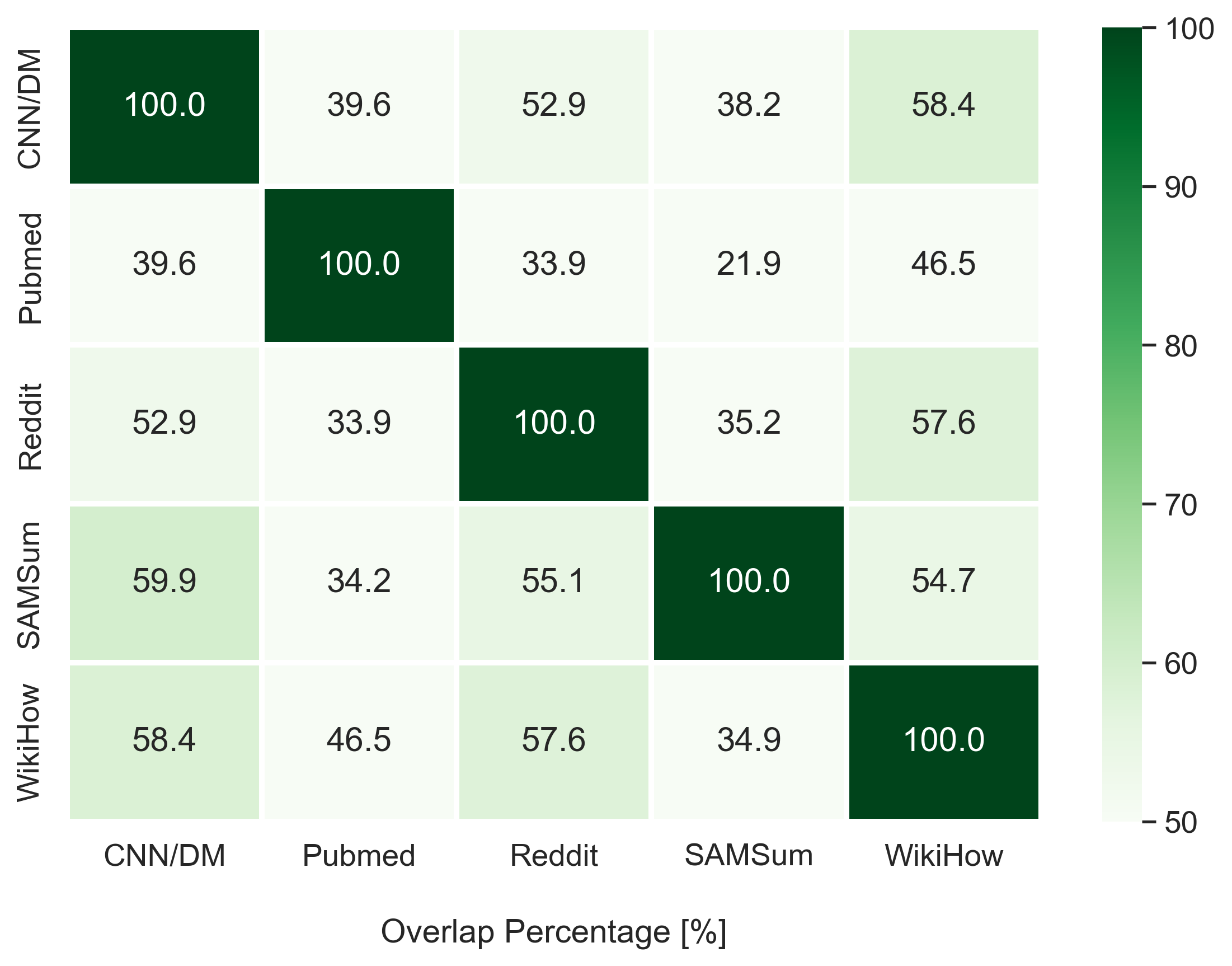}
		\subcaption{Genre Shift}
		\label{fig:Genre Shift VO}
	\end{minipage} 
	\begin{minipage}[c]{0.32\linewidth}
		\centering
		\includegraphics[width=\textwidth]{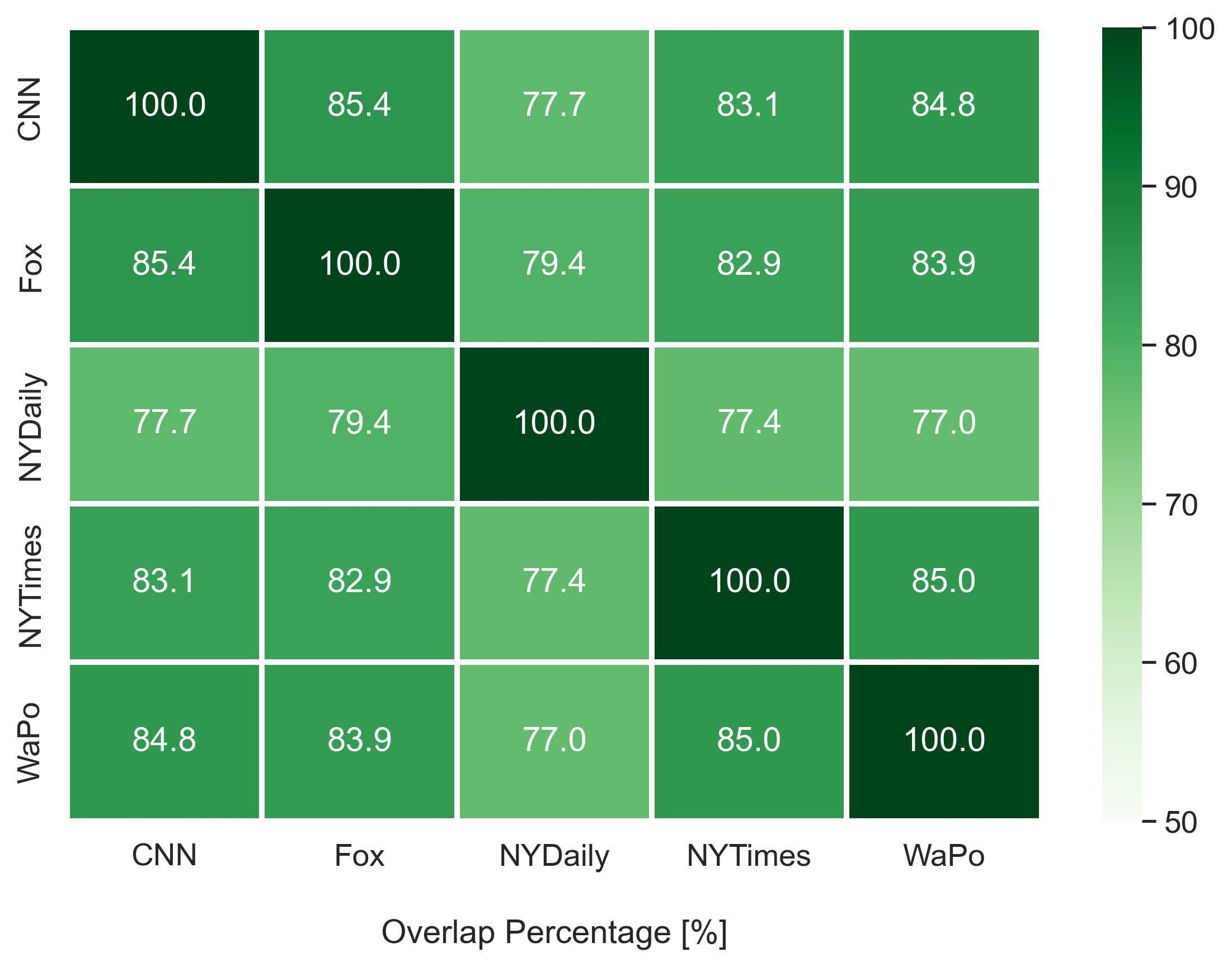}
		\subcaption{Style Shift}
		\label{fig:Style Shift VO}
	\end{minipage}
        \begin{minipage}[c]{0.32\linewidth}
		\centering
		\includegraphics[width=\textwidth]{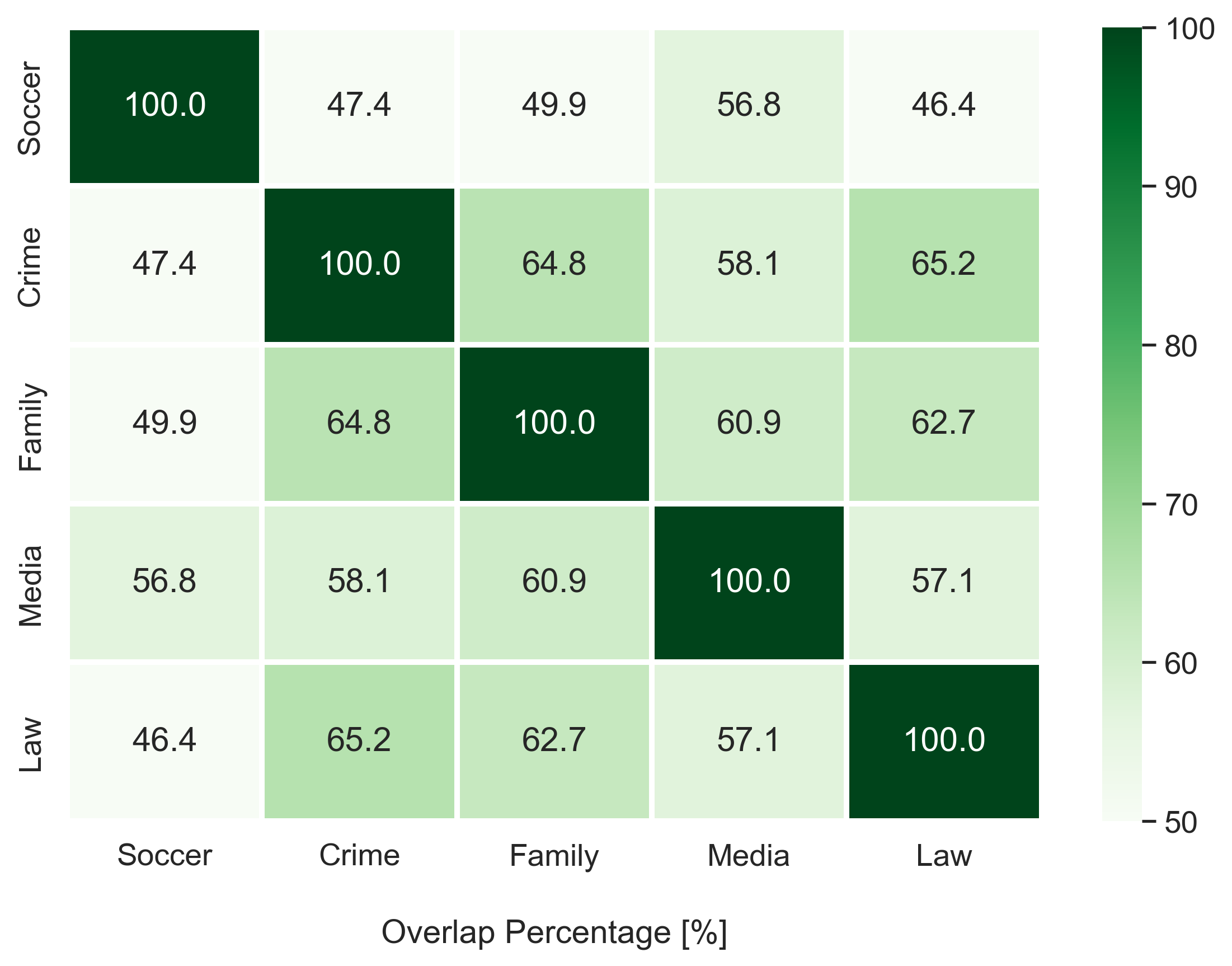}
		\subcaption{Topic Shift}
		\label{fig:Topic Shift VO}
	\end{minipage}
        \caption{Vocabulary overlaps of different levels of DomainSum . Vocabularies for each domain are created by considering the top 10K most frequent words (excluding stopwords).}
        \label{fig:VOMAP}
\end{figure*}

%% file: tables/Main-results.tex
\begin{table*}[t!]
\centering
\setlength{\tabcolsep}{3pt}

\resizebox{.99\textwidth}{!}{
\begin{tabular}{l| cc|  cc| cc| cc| cc }
    \toprule
    & ROUGE & BERTScore & ROUGE & BERTScore & ROUGE & BERTScore & ROUGE & BERTScore & ROUGE & BERTScore   \\
\midrule

    \textbf{\textsl{Genre Shift}} & \multicolumn{2}{c|}{\textbf{CNN/DM}} & \multicolumn{2}{c|}{\textbf{PubMed}} & \multicolumn{2}{c|}{\textbf{Reddit}} & \multicolumn{2}{c|}{\textbf{SAMSum}} & \multicolumn{2}{c}{\textbf{WikiHow}}   \\

    \midrule

GPT-4o mini
& \cellcolor{red!00}24.78 (-0.07) & \cellcolor{red!00}87.43 (+0.06) & \cellcolor{red!0}20.69 (+0.33) & \cellcolor{red!24}83.60 (+0.14) 
& \cellcolor{red!00}10.91 (+5.20) & \cellcolor{blue!24}81.48 (+1.58) & \cellcolor{red!00}25.01 (+4.83) & \cellcolor{red!00}89.34 (+1.08) 
& \cellcolor{red!00}15.27 (+3.39) & \cellcolor{blue!24}83.20 (+1.27)
\\
Mistral-7B
& \cellcolor{red!00}25.82 (-0.65) & \cellcolor{red!00}87.43 (-0.10) & \cellcolor{red!24}22.06 (-0.59) & \cellcolor{blue!24}83.49 (+0.28)
& \cellcolor{red!00}10.72 (+2.68) & \cellcolor{red!00}81.36 (+0.91) & \cellcolor{red!00}23.76 (+4.44) & \cellcolor{red!00}89.10 (+0.94) 
& \cellcolor{red!00}15.29 (+1.93) & \cellcolor{red!00}83.09 (+0.80) 
\\
Gemma1.1-7B
& \cellcolor{red!00}23.68 (+0.66) & \cellcolor{red!00}86.28 (+0.28) & \cellcolor{red!00}20.74 (-1.21) & \cellcolor{red!00}83.25 (+0.14) 
& \cellcolor{red!00}11.50 (-0.02) & \cellcolor{red!00}81.72 (+0.01) & \cellcolor{red!00}23.26 (+2.92) & \cellcolor{red!00}88.51 (+1.33) 
& \cellcolor{red!00}16.09 (+0.66) & \cellcolor{red!00}82.72 (+0.47) 
\\
Llama3.1-8B
& \cellcolor{red!24}26.02 (-0.80) & \cellcolor{purple!24}87.46 (+0.14) & \cellcolor{red!0}21.07 (+1.54) & \cellcolor{red!00}83.54 (+0.18)
& \cellcolor{red!00}11.83 (+3.82) & \cellcolor{red!24}81.81 (+1.07) & \cellcolor{purple!24}26.11 (+5.05) & \cellcolor{purple!24}89.61 (+0.84)
& \cellcolor{red!00}16.75 (+1.69) & \cellcolor{red!00}83.21 (+0.71)
\\
Llama3.1-70B
& \cellcolor{blue!24}25.87 (-0.13) & \cellcolor{red!00}87.11 (-2.08) & \cellcolor{blue!24}21.71 (+1.85) & \cellcolor{red!00}83.48 (-7.55)
& \cellcolor{purple!24}11.99 (+4.90) & \cellcolor{red!00}80.59 (+1.89) & \cellcolor{red!00}25.48 (+4.69) & \cellcolor{red!00}88.92 (+1.40)
& \cellcolor{purple!24}17.05 (+2.93) & \cellcolor{red!24}83.31 (+0.70)
\\

      \midrule
    \textbf{\textsl{Style Shift}} & \multicolumn{2}{c|}{\textbf{CNN}} & \multicolumn{2}{c|}{\textbf{Fox}} & \multicolumn{2}{c|}{\textbf{NYDaily}} & \multicolumn{2}{c|}{\textbf{NYTimes}} & \multicolumn{2}{c}{\textbf{WaPo}}   \\
    \midrule
        
    GPT-4o mini
    & \cellcolor{red!00}16.76 (+0.11) & \cellcolor{blue!24}84.58 (+0.11) & \cellcolor{red!0}19.06 (-0.20) & \cellcolor{red!00}85.27 (+0.17) 
    & \cellcolor{red!00}20.80 (+0.05) & \cellcolor{blue!24}85.19 (+0.34) & \cellcolor{red!00}15.37 (-0.52) & \cellcolor{red!00}84.32 (+0.16) 
    & \cellcolor{red!00}11.33 (+0.10) & \cellcolor{purple!24}83.52 (+0.18)
    \\
    Mistral-7B
    & \cellcolor{red!24}18.18 (-0.01) & \cellcolor{red!24}84.64 (-0.13) & \cellcolor{red!24}21.21 (-0.30) & \cellcolor{red!24}85.42 (+0.04)
    & \cellcolor{red!24}22.52 (-0.40) & \cellcolor{red!24}85.33 (-0.02) & \cellcolor{red!24}16.86 (+0.04) & \cellcolor{purple!24}84.45 (+0.05) 
    & \cellcolor{red!24}12.35 (+0.04) & \cellcolor{red!24}83.52 (+0.11) 
    \\
    Gemma1.1-7B
    & \cellcolor{red!00}15.94 (+0.52) & \cellcolor{red!00}82.85 (+0.34) & \cellcolor{red!00}18.92 (-0.20) & \cellcolor{red!00}84.47 (+0.47) 
    & \cellcolor{red!00}20.19 (+0.35) & \cellcolor{red!00}84.51 (-0.23) & \cellcolor{red!00}14.93 (-0.46) & \cellcolor{red!00}83.32 (+0.41) 
    & \cellcolor{red!00}10.89 (+0.22) & \cellcolor{red!00}81.51 (+0.70) 
    \\
    Llama3.1-8B
    & \cellcolor{red!00}17.46 (+1.68) & \cellcolor{red!00}83.85 (-0.08) & \cellcolor{red!00}19.95 (+2.75) & \cellcolor{blue!24}85.17 (+0.42)
    & \cellcolor{blue!24}21.30 (+2.83) & \cellcolor{red!00}85.06 (+0.37) & \cellcolor{red!00}16.19 (+0.41) & \cellcolor{red!00}84.19 (-0.03) 
    & \cellcolor{red!00}11.75 (+0.71) & \cellcolor{red!00}82.31 (+0.40)
    \\
    Llama3.1-70B
    & \cellcolor{blue!24}17.06 (+3.63) & \cellcolor{red!00}83.48 (-3.63) & \cellcolor{blue!24}19.24 (+6.96) & \cellcolor{red!00}83.30 (+2.13)
    & \cellcolor{red!0}20.24 (+3.49) & \cellcolor{red!00}81.52 (-1.22) & \cellcolor{blue!24}15.77 (+2.45) & \cellcolor{red!00}81.67 (+1.93)
    & \cellcolor{blue!24}11.60 (+1.85) & \cellcolor{red!00}78.78 (+4.09)
    \\

    \midrule
    \textbf{\textsl{Topic Shift}} & \multicolumn{2}{c|}{\textbf{Soccer}} & \multicolumn{2}{c|}{\textbf{Crime}} & \multicolumn{2}{c|}{\textbf{Family}} & \multicolumn{2}{c|}{\textbf{Media}} & \multicolumn{2}{c}{\textbf{Law}}   \\
    \midrule
    
    GPT-4o mini
    & \cellcolor{red!00}23.67 (+0.33) & \cellcolor{red!00}86.96 (+0.20)
    & \cellcolor{red!00}26.90 (+0.12) & \cellcolor{purple!24}87.92 (+0.09)
    & \cellcolor{red!00}26.85 (-0.34) & \cellcolor{purple!24}87.78 (-0.01)
    & \cellcolor{red!00}22.08 (-0.23) & \cellcolor{red!00}86.95 (+0.02)
    & \cellcolor{red!00}27.77 (+0.22) & \cellcolor{purple!24}87.95 (+0.03)
    \\
    Mistral-7B
    & \cellcolor{red!00}23.19 (+1.64) & \cellcolor{red!00}81.58 (+5.45)
    & \cellcolor{red!00}26.17 (+1.02) & \cellcolor{red!00}83.86 (+3.85)
    & \cellcolor{red!00}23.16 (+3.77) & \cellcolor{red!00}87.04 (+0.52)
    & \cellcolor{red!00}22.96 (-1.42) & \cellcolor{red!00}85.29 (-4.05)
    & \cellcolor{red!00}27.34 (+0.60) & \cellcolor{red!00}86.25 (+1.45)
    \\
    Gemma1.1-7B
    & \cellcolor{red!00}23.65 (+0.08) & \cellcolor{red!00}86.07 (+0.40)
    & \cellcolor{red!00}25.89 (-0.11) & \cellcolor{red!00}86.80 (+0.35)
    & \cellcolor{red!00}25.06 (-0.74) & \cellcolor{red!00}86.52 (+0.23)
    & \cellcolor{red!00}21.10 (-0.37) & \cellcolor{red!00}85.90 (+0.49)
    & \cellcolor{red!00}25.53 (-0.05) & \cellcolor{red!00}86.53 (+0.26)
    \\
    Llama3.1-8B
    & \cellcolor{purple!24}25.06 (+1.07) & \cellcolor{purple!24}87.01 (+0.18)
    & \cellcolor{purple!24}27.95 (+1.22) & \cellcolor{red!00}87.83 (+0.13)
    & \cellcolor{red!00}27.80 (+0.29) & \cellcolor{red!00}87.71 (-0.01)
    & \cellcolor{purple!24}23.28 (+0.60) & \cellcolor{purple!24}86.98 (+0.09)
    & \cellcolor{purple!00}28.48 (+0.75) & \cellcolor{red!00}87.86 (+0.02)
    \\
    Llama3.1-70B
    & \cellcolor{blue!00}24.53 (+1.12) & \cellcolor{blue!00}84.65 (-2.68)
    & \cellcolor{red!00}27.33 (+1.48) & \cellcolor{red!00}84.89 (+0.32)
    & \cellcolor{purple!24}28.31 (+0.45) & \cellcolor{red!00}87.10 (+0.01)
    & \cellcolor{red!00}21.73 (+2.06) & \cellcolor{red!00}83.71 (+1.14)
    & \cellcolor{purple!24}28.59 (+1.01) & \cellcolor{red!00}87.75 (-1.26)
    \\

    \bottomrule

\end{tabular}}
\caption{ Results for zero-shot and two-shot prompting across genre, style, and topic shift levels in DomainSum. Values in parentheses indicate the performance difference between two-shot and zero-shot prompting. The best results for zero-shot prompting are highlighted in red, two-shot in blue, and cases where both achieve the best results are highlighted in purple.}

\label{tab:main results}
\end{table*}

%% file: content/4experiment.tex
\section{Experiment}

\subsection{Experimental Settings}
\paragraph{Models} We investigate popular PLMs and LLMs on DomainSum. For PLMs, we consider encoder-decoder models \textbf{BART}~\cite{lewis2019bart} and \textbf{PEGASUS-X}~\cite{phang2023investigating}. For LLMs, we include OpenAI's \textbf{GPT-4o mini} \cite{openai2024gpt4technicalreport} and several top-ranked instruction-tuned open-source LLMs from Chatbot Arena \cite{zheng2023judging}\footnote{\url{https://leaderboard.lmsys.org}}, including META's \textbf{Llama3.1-8B-Instruct} and \textbf{Llama3.1-70B-Instruct} ~\cite{dubey2024llama3herdmodels}, Mistral AI's \textbf{Mistral-7B-Instruct-v0.3} ~\cite{jiang2023mistral7b}, and Google's \textbf{Gemma1.1-7B-Instruct} ~\cite{gemmateam2024gemmaopenmodelsbased}. We use the official API for GPT-4o mini\footnote{\url{https://openai.com/index/openai-api/}}, and the corresponding Hugging Face APIs\footnote{\url{https://huggingface.co/docs/api-inference/index}} for other LLMs.

We evaluated all LLMs in both zero-shot and few-shot ICL settings, and evaluate BART, PEGASUS-X, and Llama under fine-tuning settings with commonly used metrics including BERTScore F1 scores ~\cite{zhang2020bertscoreevaluatingtextgeneration} and the geometric mean of ROUGE-1/2/L scores (ROUGE) ~\cite{lin-2004-rouge}. Example zero-shot and few-shot ICL prompts can be found in Appendix~\ref{sec:prompts}. More details on the fine-tuning settings can be found in Appendix~\ref{sec:appendix setting}.

\paragraph{Data Sampling}
Due to the significant variation in the number of instances across datasets in DomainSum, we sample 10,000 instances for training, 500 for validation, and 500 for testing from the respective train, validation, and test sets of each dataset. All models are evaluated using these sampled test sets, with BART, PEGASUS-X and Llama3.1-8B fine-tuned on the corresponding sampled training sets. For LLMs, we also sample training sets of 20 and 1,000 instances for two-shot prompting and fine-tuning experiments. In the two-shot prompting setup, LLMs are prompted with two randomly selected instances from the training set. All sampled datasets and prompts used for ICL are released alongside our source code.

\subsection{In-domain Evaluation}

\paragraph{Zero/Few-Shot Results} Table~\ref{tab:main results} presents the performance of models across genre, style, and topic levels in zero-shot and two-shot settings, evaluated using ROUGE and BERTScore metrics. At the genre level, Llama3.1-8B achieves the highest performance across most datasets, while Llama3.1-70B shows more substantial improvements when transitioning from zero-shot to two-shot. In the style shift setting, Mistral-7B performs best in zero-shot, but Llama3.1-70B outperforms it in two-shot. Finally, in the topic shift setting, Llama3.1-8B consistently outperforms other models in both zero-shot and two-shot settings, with GPT-4O mini achieving the best BERTScore on three datasets. Additionally, we observe that increasing the model size from Llama3.1-8B to Llama3.1-70B does not lead to significant overall performance improvements.

\input{tables/finetune-results}

\paragraph{Fine-Tuning Results}
Table~\ref{tab:fine-tune results} presents the in-domain fine-tuning results. The results show that fine-tuned models generally outperform zero-shot and few-shot LLMs, underscoring the importance of in-domain training for this task. BART achieves the highest scores across all levels and datasets, except for slightly lower performance on PubMed. Llama3.1-8B also demonstrates significant performance improvements with fine-tuning. This is potentially due to its larger model size, which may require more data to fully benefit from in-domain training.

\paragraph{Cross-Level Results Comparison} We also compare the performance differences of models across these three levels. As shown in Table~\ref{tab:main results} and  Table~\ref{tab:fine-tune results}, in the topic shift, models perform consistently well in both zero-shot and two-shot settings, suggesting that topic shifts are easier for models to generalize. In contrast, genre shift proves to be the most challenging, with models showing significant performance variations, particularly on datasets with specific summary characteristics, such as the long summaries in PubMed. Even with two-shot prompting, improvements remain limited, reflecting the more pronounced data distribution differences in genre shifts. Style shift falls in between, with models exhibiting moderate performance variations, corresponding to the intermediate complexity of its data distribution. These findings further support the hierarchical categorization of domain shifts in abstractive summarization.

\input{figures/DomainAdaption}

\subsection{Cross-domain Evaluation}
Figure~\ref{fig:domain_shift_result} compares the cross-domain adaptation performance, where BART and Llama3.1-8B are \textbf{trained on one domain and tested on other domains at the same level}. According to the results, the genre shift (from CNN/DM) poses the greatest challenge for both models, showing significant performance drops when tested on domains outside the training set. The style shift (from CNN) presents a more moderate challenge, with models showing smaller but still noticeable performance variability. Although style-based differences are easier to manage than genre shifts, models still struggle to adapt to varied writing styles. The topic shift (from Soccer) introduces the least variability in performance, suggesting that topic-based domain changes are easier for models to handle, as they tend to share more common contextual or thematic elements.

When comparing the models, the LLM (Llama3.1-8B) does not always outperform the PLM (BART) in absolute performance scores. In several cases, particularly after fine-tuning, BART achieves higher ROUGE and BERTScore values across different domains. This suggests that PLMs like BART can be more effectively fine-tuned for specific tasks, likely due to their smaller model size and more targeted pretraining, which enables them to adapt more precisely during fine-tuning. In contrast, Llama3.1-8B demonstrates better generalization potential, particularly in style and topic shifts, where its performance remains more stable across different domains. We infer that LLMs require more extensive or tailored fine-tuning to fully leverage their capacity, as their larger scale and broader pretraining may make fine-tuning less focused and efficient compared to PLMs.

Overall, the experimental results indicate that summarization domain shifts have hierarchical structures, with genre shifts being the most challenging for models to handle, followed by style shifts, while topic shifts lead to the most stable cross-domain performance.

\input{tables/correlation}

%% file: tables/finetune-results.tex
\begin{table*}[t!]
\centering
\setlength{\tabcolsep}{3pt}

\resizebox{.99\textwidth}{!}{
\begin{tabular}{l| cc|  cc| cc| cc| cc }
    \toprule
    & ROUGE & BERTScore & ROUGE & BERTScore & ROUGE & BERTScore & ROUGE & BERTScore & ROUGE & BERTScore   \\
\midrule

    \textbf{\textsl{Genre Shift}} & \multicolumn{2}{c|}{\textbf{CNN/DM}} & \multicolumn{2}{c|}{\textbf{PubMed}} & \multicolumn{2}{c|}{\textbf{Reddit}} & \multicolumn{2}{c|}{\textbf{SAMSum}} & \multicolumn{2}{c}{\textbf{WikiHow}}   \\

    \midrule

    BART
    & \cellcolor{red!24}28.43 & \cellcolor{red!24}87.72 
    & \cellcolor{red!00}20.70 & \cellcolor{red!00}84.05 
    & \cellcolor{red!24}20.86 & \cellcolor{red!24}87.46
    & \cellcolor{red!24}37.61 & \cellcolor{red!24}91.56 
    & \cellcolor{red!24}24.73 & \cellcolor{red!24}87.62 
    \\
    PEGASUS-X
    & \cellcolor{red!00}25.54 & \cellcolor{red!00}86.73 
    & \cellcolor{red!00}21.31 & \cellcolor{red!00}83.38 
    & \cellcolor{red!00}15.74 & \cellcolor{red!00}85.37 
    & \cellcolor{red!00}31.54 & \cellcolor{red!00}90.40 
    & \cellcolor{red!00}16.75 & \cellcolor{red!00}83.08 
    \\
    Llama3.1-8B
    & \cellcolor{red!00}25.81 & \cellcolor{red!00}86.91 
    & \cellcolor{red!24}24.10 & \cellcolor{red!24}84.44 
    & \cellcolor{red!00}13.31 & \cellcolor{red!00}79.69 
    & \cellcolor{red!00}17.67 & \cellcolor{red!00}87.29 
    & \cellcolor{red!00}19.78 & \cellcolor{red!00}83.53 
    \\

      \midrule
    \textbf{\textsl{Style Shift}} & \multicolumn{2}{c|}{\textbf{CNN}} & \multicolumn{2}{c|}{\textbf{Fox}} & \multicolumn{2}{c|}{\textbf{NYDaily}} & \multicolumn{2}{c|}{\textbf{NYTimes}} & \multicolumn{2}{c}{\textbf{WaPo}}   \\
    \midrule
        
        BART
    & \cellcolor{red!24}42.02 & \cellcolor{red!24}89.44 
    & \cellcolor{red!24}55.05 & \cellcolor{red!24}91.55 
    & \cellcolor{red!24}53.84 & \cellcolor{red!24}91.06 
    & \cellcolor{red!24}25.38 & \cellcolor{red!24}86.81 
    & \cellcolor{red!24}20.05 & \cellcolor{red!24}85.58 
    \\
    PEGASUS-X
    & \cellcolor{red!00}38.20 & \cellcolor{red!00}88.16 
    & \cellcolor{red!00}54.58 & \cellcolor{red!00}91.34 
    & \cellcolor{red!00}49.45 & \cellcolor{red!00}90.33 
    & \cellcolor{red!00}23.62 & \cellcolor{red!00}85.82 
    & \cellcolor{red!00}18.34 & \cellcolor{red!00}84.39 
    \\
    Llama3.1-8B
    & \cellcolor{red!00}20.55 & \cellcolor{red!00}85.27 
    & \cellcolor{red!00}29.20 & \cellcolor{red!00}86.84 
    & \cellcolor{red!00}33.91 & \cellcolor{red!00}87.70 
    & \cellcolor{red!00}16.33 & \cellcolor{red!00}83.60 
    & \cellcolor{red!00}10.66 & \cellcolor{red!00}82.70 
    \\

    \midrule
    \textbf{\textsl{Topic Shift}} & \multicolumn{2}{c|}{\textbf{Soccer}} & \multicolumn{2}{c|}{\textbf{Crime}} & \multicolumn{2}{c|}{\textbf{Family}} & \multicolumn{2}{c|}{\textbf{Media}} & \multicolumn{2}{c}{\textbf{Law}}   \\
    \midrule
    
       BART
    & \cellcolor{red!24}29.71 & \cellcolor{red!24}88.90 
    & \cellcolor{red!24}31.45 & \cellcolor{red!24}88.98
    & \cellcolor{red!24}30.45 & \cellcolor{red!24}88.67 
    & \cellcolor{red!24}27.56 & \cellcolor{red!24}88.20 
    & \cellcolor{red!24}31.50 & \cellcolor{red!24}88.86 
    \\
    PEGASUS-X
    & \cellcolor{red!00}24.74 & \cellcolor{red!00}87.79 
    & \cellcolor{red!00}27.32 & \cellcolor{red!00}88.09 
    & \cellcolor{red!00}27.00 & \cellcolor{red!00}87.89
    & \cellcolor{red!00}24.12 & \cellcolor{red!00}87.54 
    & \cellcolor{red!00}28.17 & \cellcolor{red!00}88.19 
    \\
    Llama3.1-8B
    & \cellcolor{red!00}23.13 & \cellcolor{red!00}85.98 
    & \cellcolor{red!00}28.18 & \cellcolor{red!00}87.39 
    & \cellcolor{red!00}24.35 & \cellcolor{red!00}85.72 
    & \cellcolor{red!00}23.11 & \cellcolor{red!00}85.71 
    & \cellcolor{red!00}27.39 & \cellcolor{red!00}87.10 
    \\

    \bottomrule

\end{tabular}}
\caption{Fine-tuning performance across five domains based on genre, style, and topic shift. Models are trained and evaluated within the same domain.}
\label{tab:fine-tune results}
\end{table*}

%% file: figures/DomainAdaption.tex
\begin{figure}[t!]
    \centering
    \includegraphics[width=0.98\linewidth]{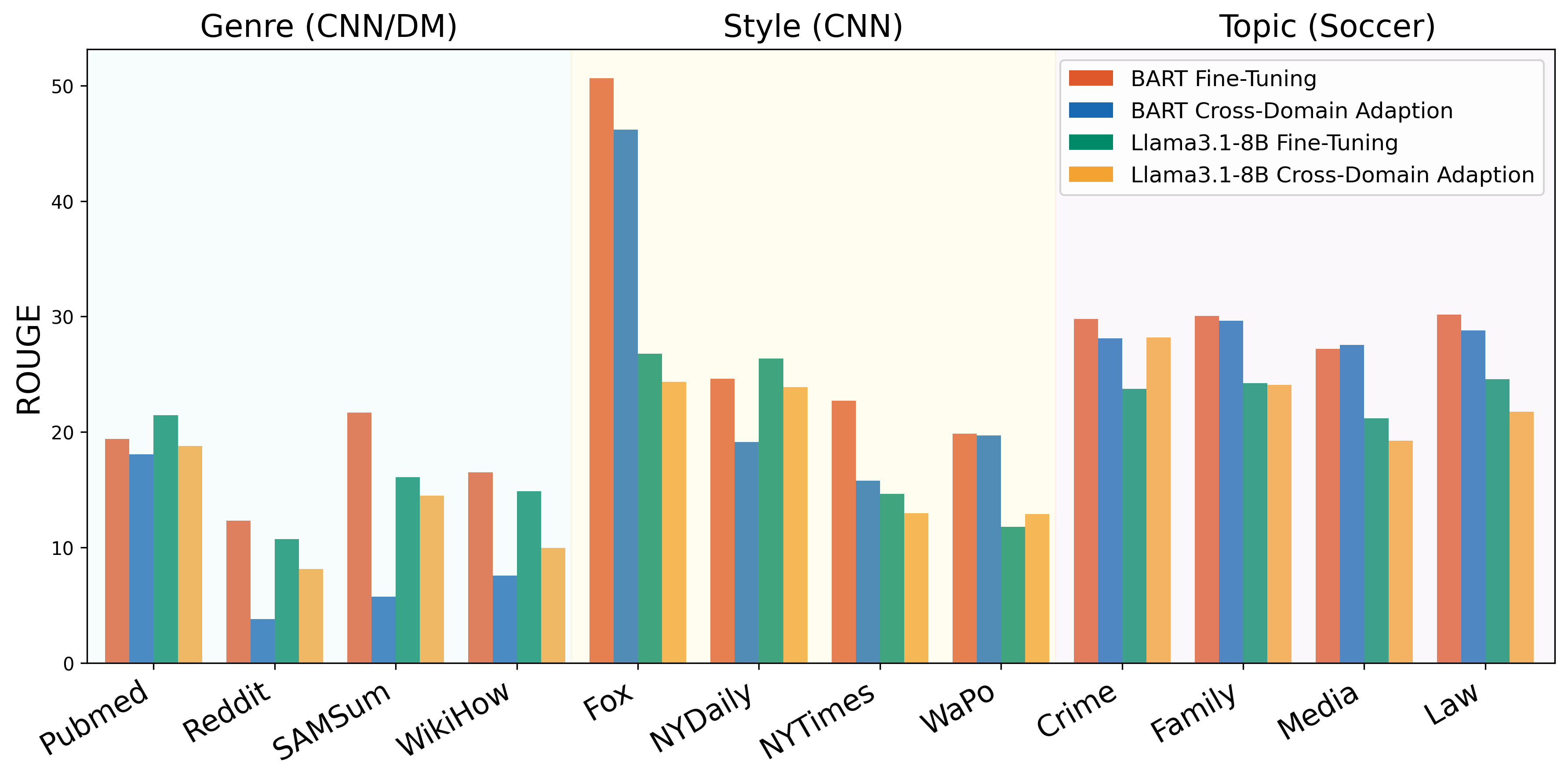}
    \caption{Comparison of in-domain fine-tuning and domain adaptation results for BART and LLaMA-3.1-8B models. The x-axis represents the target domains, with region colors indicating the source domains.}
    \label{fig:domain_shift_result}
\end{figure}

%% file: tables/correlation.tex
\begin{table}[ht]
\centering
\small
\begin{tabular}{lccc}
\toprule
Measure & Genre & Style & Topic \\ 
\midrule
Doc Length        & -0.305 & -0.935* & 0.275 \\
Sum Length        & -0.159 & 0.436 & 0.886* \\
Compression       & -0.431 & -0.915* & -0.680 \\
Doc Diversity     & 0.022  & 0.944*  & -0.617 \\
Sum Diversity     & -0.319 & -0.998* & -0.929* \\
Coverage          & 0.313  & 0.994*  & 0.764 \\
Abstractiveness   & -0.321 & -0.997* & -0.718 \\
Density           & 0.424  & 0.830  & 0.155 \\
\bottomrule
\end{tabular}
\caption{Pearson correlation coefficients between domain characteristic measures and model performance, averaged across model performance under three settings: zero-shot prompting, two-shot prompting, and fine-tuning. Asterisks (*) denote statistically significant correlations ($p \leq 0.05$).}
\label{tab:correlations}
\end{table}

%% file: content/5analysis.tex
\subsection{Correlation Analysis}
We present the Pearson correlation coefficients between our summarization measures and model performance across the genre, style, and topic dataset levels. As shown in Table~\ref{tab:correlations}, the correlation patterns differ significantly across these levels. At the genre level, none of the measures exhibit statistically significant correlations with model performance. In contrast, at the style level, several measures show significant positive or negative correlations ($p\leq0.05$). Notably, summary diversity shows the strongest negative correlation and  Coverage shows the strongest positive correlation with model performance. At the topic level, two measures exhibit significant correlations: summary length is positively correlated with performance, while summary diversity shows a significant negative correlation. The distinct correlation patterns across genre, style, and topic levels demonstrate that summarization performance is influenced by different aspects of the data distribution, and underscore the need for further investigation into how different summarization measures can be tailored to improve model performance across various domain shifts.

%% file: content/6appendix.tex
\newpage

\section{Detailed Domain Characteristic Measure Results}
\label{sec:appendix measure}
The detailed measurement results in terms of all datasets included in DomainSum is presented in Table~\ref{tab:measures}.
\input{tables/measures}

\section{Prompts}
\label{sec:prompts}
\paragraph{zero-shot prompt}

\begin{center}
\fcolorbox{black}{gray!10}{
\parbox{0.9\linewidth}{
\small 
You are an expert at summarization. Summarize the following text:  \{document\} \\

 Summary: 
}}
\end{center}

\paragraph{two-shot prompt}

\begin{center}
\fcolorbox{black}{gray!10}{
\parbox{0.9\linewidth}{
\small 

  You are an expert at summarization. Here are two examples of how to summarize a text:\\

            Example 1: \\
            Document: \{doc\_example1\} \\
            Summary: \{sum\_example1\} \\

            Example 2: \\
            Document: \{doc\_example2\} \\
            Summary: \{sum\_example2\} \\

            Now, summarize the following text: \\

            Document: \{document\} \\

            Summary: \\
 
}
}
\end{center}

The prompt used to fine-tune the Llama3.1-8B model.

\begin{center}
\fcolorbox{black}{gray!10}{
\parbox{0.9\linewidth}{
\small
\textless  \textbar im\_start\textbar \textgreater user \\
You are an expert at summarization. Summarize the following text:  \{document\} \\

 Summary:\textless  \textbar im\_end\textbar \textgreater \\
\textless  \textbar im\_start\textbar \textgreater assistant 
}}
\end{center}

\section{Detailed Fine-Tunning Settings}
\label{sec:appendix setting}

We fine-tune pre-trained BART and PEGASUS-X models using their base versions with huggingface packages~\cite{wolf-etal-2020-transformers}. We use AdamW optimizer~\cite{Loshchilov2017DecoupledWD} with learning rate set to $5 \times 10^{-5}$. We set the batch size to 12 and the number of training epoch to 4.  The best model is selected based on the development set results. For Llama3.1-8B, we use the Low Rank Adaptation (LoRA) ~\cite{hu2021loralowrankadaptationlarge} fine-tuning method with huggingface packages, where the model is loaded to GPU as quantized 8-bit weights.  We set the learning rate to $5 \times 10^{-4}$, the batch size to 4, and the number of training epoch to 20. All experiments are conducted on a single NVIDIA A100-SXM4-40GB GPU.



\section{Experimental Results}
\label{sec:domain results}
The detailed ROUGE scores and BERTScores of BART and Llama3.1-8B for cross domain adaptation on three levels in DomainSum is shown in Table~\ref{tab:adaption-bart-llama}.
\input{tables/domain_adaption}

%% file: tables/measures.tex
\begin{table*}[t]
\centering
\scriptsize  
\begin{tabular}{l|ccc|ccccc}
\hline
& \multicolumn{3}{c|}{\textbf{Measures }} & \multicolumn{5}{c}{\textbf{Measures (\%)}} \\
\hline
\textbf{Datasets} & \textbf{Doc Length} & \textbf{Sum Length} & \textbf{Compression} & \textbf{Doc Diversity} & \textbf{Sum Diversity} & \textbf{Coverage} & \textbf{Abstractiveness} & \textbf{Density} \\
\hline

\multicolumn{9}{c}{\textbf{\textsl{Genre Shift}}} \\
\hline
CNN/DM   & 772.49 & 57.87 & 14.49 & 72.53 & 92.91 & 55.24 & 70.24 & 83.81 \\
PubMed   & 444.00 & 209.51 & 2.32  & 78.86 & 80.50 & 27.55 & 87.80 & 19.44 \\
Reddit   & 483.41 & 28.00 & 18.75 & 78.30 & 91.70 & 33.72 & 88.09 & 13.44 \\
SamSum   & 126.59 & 23.09 & 5.82  & 80.34 & 95.58 & 31.62 & 90.52 & 13.85 \\
WikiHow  & 582.18 & 62.19 & 11.27 & 74.58 & 85.76 & 35.98 & 89.68 & 9.31  \\
\hline
\multicolumn{9}{c}{\textbf{\textsl{Style Shift}}} \\
\hline
CNN      & 971.18 & 30.92 & 36.91 & 75.00 & 96.40 & 63.42 & 52.14 & 10.77 \\
Fox      & 654.45 & 28.69 & 33.47 & 77.80 & 96.77 & 75.34 & 35.11 & 14.50 \\
NYDaily  & 577.10 & 34.97 & 20.65 & 78.71 & 95.96 & 78.32 & 32.78 & 13.80 \\
NYTimes  & 990.96 & 35.97 & 41.38 & 75.43 & 95.69 & 57.22 & 63.16 & 12.89 \\
WaPo     & 1069.26 & 22.89 & 60.09 & 73.08 & 97.09 & 46.94 & 73.50 & 4.21  \\
\hline
\multicolumn{9}{c}{\textbf{\textsl{Topic Shift}}} \\
\hline
Soccer   & 642.51 & 52.81 & 12.61 & 75.96 & 93.31 & 48.35 & 78.45 & 30.65 \\
Crime    & 730.59 & 66.30 & 12.11 & 71.31 & 91.97 & 58.07 & 67.37 & 94.20 \\
Family   & 870.14 & 66.64 & 14.05 & 70.76 & 92.00 & 57.66 & 68.64 & 80.25 \\
Media    & 796.58 & 56.81 & 15.60 & 73.07 & 93.22 & 53.19 & 72.12 & 63.25 \\
Law      & 796.11 & 70.50 & 12.55 & 71.77 & 91.87 & 58.83 & 66.48 & 84.47 \\
\hline
\end{tabular}

\caption{Average Measures of domain adaptation summarization datasets. Length is represented in average tokens, Compression is a ratio, and diversity, coverage, abstractiveness, and density are expressed as percentages.}
\label{tab:measures}
\end{table*}

%% file: tables/domain_adaption.tex
\begin{table}[t!]
\centering
\footnotesize
\resizebox{.96\linewidth}{!}{
\begin{tabular}{lcccc}

\toprule
Model & \makecell{Training \\ Domain} & \makecell{Test \\ Domain} & \makecell{ROUGE} & \makecell{BERTScore} \\
\midrule

\multirow{12}{*}{BART} & \multirow{4}{*}{CNN/DM}  & Pubmed   & 19.39 ($\downarrow$1.31) & 84.03 ($\downarrow$0.02) \\
                       &                          & Reddit   & 12.33 ($\downarrow$8.53) & 84.23 ($\downarrow$3.23) \\
                       &                          & SAMSum   & 21.68 ($\downarrow$15.93) & 86.61 ($\downarrow$4.95) \\
                       &                          & WikiHow  & 16.39 ($\downarrow$8.34) & 85.07 ($\downarrow$2.55) \\ 
\cline{2-5} 
                       & \multirow{4}{*}{CNN}     & Fox      & 50.63 ($\downarrow$4.42) & 90.98 ($\uparrow$0.57) \\
                       &                          & NYDaily  & 47.53 ($\downarrow$6.31) & 90.18 ($\downarrow$0.88) \\
                       &                          & NYTimes  & 22.69 ($\downarrow$2.69) & 86.30 ($\downarrow$0.51) \\
                       &                          & WaPo     & 19.87 ($\downarrow$0.18) & 85.52 ($\uparrow$0.06) \\
\cline{2-5}
                       & \multirow{4}{*}{Soccer} & Crime    & 29.78 ($\downarrow$1.67) & 88.67 ($\downarrow$0.31) \\
                       &   
                       & Family   & 30.05 ($\downarrow$0.40) & 88.78 ($\uparrow$0.11) \\

                       &                          & Media    & 27.20 ($\uparrow$0.36) & 88.21 ($\uparrow$0.01) \\
                       &                          & Law      & 30.15 ($\downarrow$1.35) & 88.77 ($\downarrow$0.09) \\

\midrule
\multirow{12}{*}{Llama3.1-8B} & \multirow{4}{*}{CNN/DM}  & Pubmed   & 21.45 ($\downarrow$2.65) & 82.43 ($\downarrow$2.01) \\
                              &                          & Reddit   & 10.73 ($\downarrow$2.58) & 80.50 ($\uparrow$0.81) \\
                              &                          & SAMSum   & 16.07 ($\downarrow$1.60) & 86.39 ($\downarrow$0.90) \\
                              &                          & WikiHow  & 14.87 ($\downarrow$4.91) & 82.56 ($\downarrow$0.97) \\
\cline{2-5}
                              & \multirow{4}{*}{CNN}     & Fox      & 26.77 ($\downarrow$2.43) & 86.48 ($\downarrow$0.36) \\
                              &                          & NYDaily  & 26.36 ($\downarrow$7.55) & 86.26 ($\downarrow$1.44) \\
                              &                          & NYTimes  & 14.64 ($\downarrow$1.69) & 83.93 ($\uparrow$0.33) \\
                              &                          & WaPo     & 11.78 ($\uparrow$1.12) & 83.19 ($\uparrow$0.49) \\
\cline{2-5}
                              & \multirow{4}{*}{Soccer} & Crime    & 23.73 ($\downarrow$4.45) & 86.32 ($\downarrow$1.07) \\
                              & & Family   & 24.22 ($\downarrow$0.13) & 86.55 ($\uparrow$0.83) \\
                              
                              &                          & Media    & 21.18 ($\downarrow$1.93) & 86.04 ($\uparrow$0.34) \\
                              &                          & Law      & 24.58 ($\downarrow$2.81) & 86.53 ($\downarrow$0.57) \\
\bottomrule

\end{tabular}}
\caption{Results of single-domain adaptation based on fine-tuned BART and Llama3.1-8B models. The numbers in parentheses represent the performance gain for cross-domain training compared to fine-tuning; a negative value indicates a performance decrease in cross-domain training.}
\label{tab:adaption-bart-llama}
\end{table}

%% file: main.bbl
\begin{thebibliography}{52}
\providecommand{\natexlab}[1]{#1}

\bibitem[{Afzal et~al.(2024)Afzal, Chalumattu, Matthes, and Espuny}]{afzal2024adapteval}
Anum Afzal, Ribin Chalumattu, Florian Matthes, and Laura~Mascarell Espuny. 2024.
\newblock Adapteval: Evaluating large language models on domain adaptation for text summarization.
\newblock \emph{arXiv preprint arXiv:2407.11591}.

\bibitem[{Blei et~al.(2003)Blei, Ng, and Jordan}]{blei2003latent}
David~M Blei, Andrew~Y Ng, and Michael~I Jordan. 2003.
\newblock Latent dirichlet allocation.
\newblock \emph{Journal of machine Learning research}, 3(Jan):993--1022.

\bibitem[{Brown(2020)}]{brown2020language}
Tom~B Brown. 2020.
\newblock Language models are few-shot learners.
\newblock \emph{arXiv preprint arXiv:2005.14165}.

\bibitem[{Cao et~al.(2017)Cao, Li, Li, and Wei}]{cao2017improving}
Ziqiang Cao, Wenjie Li, Sujian Li, and Furu Wei. 2017.
\newblock Improving multi-document summarization via text classification.
\newblock In \emph{Proceedings of the AAAI conference on artificial intelligence}, 1.

\bibitem[{Cohan et~al.(2018)Cohan, Dernoncourt, Kim, Bui, Kim, Chang, and Goharian}]{cohan2018discourse}
Arman Cohan, Franck Dernoncourt, Doo~Soon Kim, Trung Bui, Seokhwan Kim, Walter Chang, and Nazli Goharian. 2018.
\newblock A discourse-aware attention model for abstractive summarization of long documents.
\newblock \emph{arXiv preprint arXiv:1804.05685}.

\bibitem[{DiMarco and Hirst(1993)}]{dimarco1993computational}
Chrysanne DiMarco and Graeme Hirst. 1993.
\newblock A computational theory of goal-directed style in syntax.
\newblock \emph{Computational Linguistics}, 19(3):451--500.

\bibitem[{Dubey et~al.(2024)Dubey, Jauhri, Pandey et~al.}]{dubey2024llama3herdmodels}
Abhimanyu Dubey, Abhinav Jauhri, Abhinav Pandey, et~al. 2024.
\newblock \href {https://arxiv.org/abs/2407.21783} {The llama 3 herd of models}.
\newblock \emph{Preprint}, arXiv:2407.21783.

\bibitem[{Fabbri et~al.(2020)Fabbri, Han, Li, Li, Ghazvininejad, Joty, Radev, and Mehdad}]{fabbri2020improving}
Alexander~R Fabbri, Simeng Han, Haoyuan Li, Haoran Li, Marjan Ghazvininejad, Shafiq Joty, Dragomir Radev, and Yashar Mehdad. 2020.
\newblock Improving zero and few-shot abstractive summarization with intermediate fine-tuning and data augmentation.
\newblock \emph{arXiv preprint arXiv:2010.12836}.

\bibitem[{Ganin et~al.(2016)Ganin, Ustinova, Ajakan, Germain, Larochelle, Laviolette, March, and Lempitsky}]{ganin2016domain}
Yaroslav Ganin, Evgeniya Ustinova, Hana Ajakan, Pascal Germain, Hugo Larochelle, Fran{\c{c}}ois Laviolette, Mario March, and Victor Lempitsky. 2016.
\newblock Domain-adversarial training of neural networks.
\newblock \emph{Journal of machine learning research}, 17(59):1--35.

\bibitem[{Gliwa et~al.(2019)Gliwa, Mochol, Biesek, and Wawer}]{gliwa2019samsum}
Bogdan Gliwa, Iwona Mochol, Maciej Biesek, and Aleksander Wawer. 2019.
\newblock Samsum corpus: A human-annotated dialogue dataset for abstractive summarization.
\newblock \emph{arXiv preprint arXiv:1911.12237}.

\bibitem[{Goyal et~al.(2022)Goyal, Li, and Durrett}]{goyal2022news}
Tanya Goyal, Junyi~Jessy Li, and Greg Durrett. 2022.
\newblock News summarization and evaluation in the era of gpt-3.
\newblock \emph{arXiv preprint arXiv:2209.12356}.

\bibitem[{Grusky et~al.(2018)Grusky, Naaman, and Artzi}]{grusky2018newsroom}
Max Grusky, Mor Naaman, and Yoav Artzi. 2018.
\newblock Newsroom: A dataset of 1.3 million summaries with diverse extractive strategies.
\newblock In \emph{Proceedings of the 2018 Conference of the North American Chapter of the Association for Computational Linguistics: Human Language Technologies, Volume 1 (Long Papers)}, pages 708--719.

\bibitem[{Gururangan et~al.(2020)Gururangan, Marasovi{\'c}, Swayamdipta, Lo, Beltagy, Downey, and Smith}]{gururangan2020don}
Suchin Gururangan, Ana Marasovi{\'c}, Swabha Swayamdipta, Kyle Lo, Iz~Beltagy, Doug Downey, and Noah~A Smith. 2020.
\newblock Don't stop pretraining: Adapt language models to domains and tasks.
\newblock \emph{arXiv preprint arXiv:2004.10964}.

\bibitem[{Hermann et~al.(2015)Hermann, Kocisky, Grefenstette, Espeholt, Kay, Suleyman, and Blunsom}]{hermann2015teaching}
Karl~Moritz Hermann, Tomas Kocisky, Edward Grefenstette, Lasse Espeholt, Will Kay, Mustafa Suleyman, and Phil Blunsom. 2015.
\newblock Teaching machines to read and comprehend.
\newblock \emph{Advances in neural information processing systems}, 28.

\bibitem[{Hu et~al.(2021)Hu, Shen, Wallis, Allen-Zhu, Li, Wang, Wang, and Chen}]{hu2021loralowrankadaptationlarge}
Edward~J. Hu, Yelong Shen, Phillip Wallis, Zeyuan Allen-Zhu, Yuanzhi Li, Shean Wang, Lu~Wang, and Weizhu Chen. 2021.
\newblock \href {https://arxiv.org/abs/2106.09685} {Lora: Low-rank adaptation of large language models}.
\newblock \emph{Preprint}, arXiv:2106.09685.

\bibitem[{Hua and Wang(2017)}]{hua2017pilot}
Xinyu Hua and Lu~Wang. 2017.
\newblock A pilot study of domain adaptation effect for neural abstractive summarization.
\newblock \emph{arXiv preprint arXiv:1707.07062}.

\bibitem[{Jiang et~al.(2023)Jiang, Sablayrolles, Mensch et~al.}]{jiang2023mistral7b}
Albert~Q. Jiang, Alexandre Sablayrolles, Arthur Mensch, et~al. 2023.
\newblock \href {https://arxiv.org/abs/2310.06825} {Mistral 7b}.
\newblock \emph{Preprint}, arXiv:2310.06825.

\bibitem[{Joshi et~al.(2012)Joshi, Dredze, Cohen, and Rose}]{joshi2012multi}
Mahesh Joshi, Mark Dredze, William Cohen, and Carolyn Rose. 2012.
\newblock Multi-domain learning: when do domains matter?
\newblock In \emph{Proceedings of the 2012 Joint Conference on Empirical Methods in Natural Language Processing and Computational Natural Language Learning}, pages 1302--1312.

\bibitem[{Kim et~al.(2018)Kim, Kim, and Kim}]{kim2018abstractive}
Byeongchang Kim, Hyunwoo Kim, and Gunhee Kim. 2018.
\newblock Abstractive summarization of reddit posts with multi-level memory networks.
\newblock \emph{arXiv preprint arXiv:1811.00783}.

\bibitem[{Koupaee and Wang(2018)}]{koupaee2018wikihow}
Mahnaz Koupaee and William~Yang Wang. 2018.
\newblock Wikihow: A large scale text summarization dataset.
\newblock \emph{arXiv preprint arXiv:1810.09305}.

\bibitem[{Laskar et~al.(2022)Laskar, Hoque, and Huang}]{laskar2022domain}
Md~Tahmid~Rahman Laskar, Enamul Hoque, and Jimmy~Xiangji Huang. 2022.
\newblock Domain adaptation with pre-trained transformers for query-focused abstractive text summarization.
\newblock \emph{Computational Linguistics}, 48(2):279--320.

\bibitem[{Lewis(2019)}]{lewis2019bart}
M~Lewis. 2019.
\newblock Bart: Denoising sequence-to-sequence pre-training for natural language generation, translation, and comprehension.
\newblock \emph{arXiv preprint arXiv:1910.13461}.

\bibitem[{Li et~al.(2024)Li, Miao, Huang, and Gao}]{li2024word}
Yinghao Li, Siyu Miao, Heyan Huang, and Yang Gao. 2024.
\newblock Word matters: What influences domain adaptation in summarization?
\newblock \emph{arXiv preprint arXiv:2406.14828}.

\bibitem[{Lin(2004)}]{lin-2004-rouge}
Chin-Yew Lin. 2004.
\newblock \href {https://aclanthology.org/W04-1013} {{ROUGE}: A package for automatic evaluation of summaries}.
\newblock In \emph{Text Summarization Branches Out}, pages 74--81, Barcelona, Spain. Association for Computational Linguistics.

\bibitem[{Liu and Lapata(2019)}]{liu2019text}
Yang Liu and Mirella Lapata. 2019.
\newblock Text summarization with pretrained encoders.
\newblock \emph{arXiv preprint arXiv:1908.08345}.

\bibitem[{Liu et~al.(2022)Liu, Liu, Radev, and Neubig}]{liu2022brio}
Yixin Liu, Pengfei Liu, Dragomir Radev, and Graham Neubig. 2022.
\newblock Brio: Bringing order to abstractive summarization.
\newblock \emph{arXiv preprint arXiv:2203.16804}.

\bibitem[{Loshchilov and Hutter(2017)}]{Loshchilov2017DecoupledWD}
Ilya Loshchilov and Frank Hutter. 2017.
\newblock \href {https://api.semanticscholar.org/CorpusID:53592270} {Decoupled weight decay regularization}.
\newblock In \emph{International Conference on Learning Representations}.

\bibitem[{Magooda and Litman(2020)}]{magooda2020abstractive}
Ahmed Magooda and Diane Litman. 2020.
\newblock Abstractive summarization for low resource data using domain transfer and data synthesis.
\newblock In \emph{The thirty-third international flairs conference}.

\bibitem[{Min et~al.(2022)Min, Lyu, Holtzman, Artetxe, Lewis, Hajishirzi, and Zettlemoyer}]{min2022rethinking}
Sewon Min, Xinxi Lyu, Ari Holtzman, Mikel Artetxe, Mike Lewis, Hannaneh Hajishirzi, and Luke Zettlemoyer. 2022.
\newblock Rethinking the role of demonstrations: What makes in-context learning work?
\newblock \emph{arXiv preprint arXiv:2202.12837}.

\bibitem[{Narayan et~al.(2018)Narayan, Cohen, and Lapata}]{narayan2018don}
Shashi Narayan, Shay~B Cohen, and Mirella Lapata. 2018.
\newblock Don’t give me the details, just the summary! topic-aware convolutional neural networks for extreme summarization.
\newblock In \emph{Proceedings of the 2018 Conference on Empirical Methods in Natural Language Processing}, pages 1797--1807.

\bibitem[{OpenAI et~al.(2024)OpenAI, Achiam, Adler et~al.}]{openai2024gpt4technicalreport}
OpenAI, Josh Achiam, Steven Adler, et~al. 2024.
\newblock \href {https://arxiv.org/abs/2303.08774} {Gpt-4 technical report}.
\newblock \emph{Preprint}, arXiv:2303.08774.

\bibitem[{Phang et~al.(2023)Phang, Zhao, and Liu}]{phang2023investigating}
Jason Phang, Yao Zhao, and Peter~J Liu. 2023.
\newblock Investigating efficiently extending transformers for long input summarization.
\newblock In \emph{Proceedings of the 2023 Conference on Empirical Methods in Natural Language Processing}, pages 3946--3961.

\bibitem[{Ramponi and Plank(2020)}]{ramponi2020neural}
Alan Ramponi and Barbara Plank. 2020.
\newblock Neural unsupervised domain adaptation in nlp---a survey.
\newblock \emph{arXiv preprint arXiv:2006.00632}.

\bibitem[{Sutskever(2014)}]{sutskever2014sequence}
I~Sutskever. 2014.
\newblock Sequence to sequence learning with neural networks.
\newblock \emph{arXiv preprint arXiv:1409.3215}.

\bibitem[{Team et~al.(2024)Team, Mesnard, Hardin et~al.}]{gemmateam2024gemmaopenmodelsbased}
Gemma Team, Thomas Mesnard, Cassidy Hardin, et~al. 2024.
\newblock \href {https://arxiv.org/abs/2403.08295} {Gemma: Open models based on gemini research and technology}.
\newblock \emph{Preprint}, arXiv:2403.08295.

\bibitem[{Wang et~al.(2019)Wang, Liu, Zhong, Fu, Qiu, and Huang}]{wang2019exploring}
Danqing Wang, Pengfei Liu, Ming Zhong, Jie Fu, Xipeng Qiu, and Xuanjing Huang. 2019.
\newblock Exploring domain shift in extractive text summarization.
\newblock \emph{arXiv preprint arXiv:1908.11664}.

\bibitem[{Wang et~al.(2023)Wang, Zhang, and Wang}]{wang2023element}
Yiming Wang, Zhuosheng Zhang, and Rui Wang. 2023.
\newblock Element-aware summarization with large language models: Expert-aligned evaluation and chain-of-thought method.
\newblock \emph{arXiv preprint arXiv:2305.13412}.

\bibitem[{Wei et~al.(2022)Wei, Wang, Schuurmans, Bosma, Xia, Chi, Le, Zhou et~al.}]{wei2022chain}
Jason Wei, Xuezhi Wang, Dale Schuurmans, Maarten Bosma, Fei Xia, Ed~Chi, Quoc~V Le, Denny Zhou, et~al. 2022.
\newblock Chain-of-thought prompting elicits reasoning in large language models.
\newblock \emph{Advances in neural information processing systems}, 35:24824--24837.

\bibitem[{Wolf et~al.(2020)Wolf, Debut, Sanh, Chaumond, Delangue, Moi, Cistac, Rault, Louf, Funtowicz, Davison, Shleifer, von Platen, Ma, Jernite, Plu, Xu, Le~Scao, Gugger, Drame, Lhoest, and Rush}]{wolf-etal-2020-transformers}
Thomas Wolf, Lysandre Debut, Victor Sanh, Julien Chaumond, Clement Delangue, Anthony Moi, Pierric Cistac, Tim Rault, Remi Louf, Morgan Funtowicz, Joe Davison, Sam Shleifer, Patrick von Platen, Clara Ma, Yacine Jernite, Julien Plu, Canwen Xu, Teven Le~Scao, Sylvain Gugger, Mariama Drame, Quentin Lhoest, and Alexander Rush. 2020.
\newblock \href {https://doi.org/10.18653/v1/2020.emnlp-demos.6} {Transformers: State-of-the-art natural language processing}.
\newblock In \emph{Proceedings of the 2020 Conference on Empirical Methods in Natural Language Processing: System Demonstrations}, pages 38--45, Online. Association for Computational Linguistics.

\bibitem[{Yogatama et~al.(2015)Yogatama, Liu, and Smith}]{yogatama2015extractive}
Dani Yogatama, Fei Liu, and Noah~A Smith. 2015.
\newblock Extractive summarization by maximizing semantic volume.
\newblock In \emph{Proceedings of the 2015 Conference on Empirical Methods in Natural Language Processing}, pages 1961--1966.

\bibitem[{Yu et~al.(2021)Yu, Liu, and Fung}]{yu2021adaptsum}
Tiezheng Yu, Zihan Liu, and Pascale Fung. 2021.
\newblock Adaptsum: Towards low-resource domain adaptation for abstractive summarization.
\newblock \emph{arXiv preprint arXiv:2103.11332}.

\bibitem[{Zhang et~al.(2022)Zhang, Liu, and Zhang}]{zhang2022hegel}
Haopeng Zhang, Xiao Liu, and Jiawei Zhang. 2022.
\newblock Hegel: Hypergraph transformer for long document summarization.
\newblock \emph{arXiv preprint arXiv:2210.04126}.

\bibitem[{Zhang et~al.(2023{\natexlab{a}})Zhang, Liu, and Zhang}]{zhang2023contrastive}
Haopeng Zhang, Xiao Liu, and Jiawei Zhang. 2023{\natexlab{a}}.
\newblock Contrastive hierarchical discourse graph for scientific document summarization.
\newblock In \emph{The 4th workshop on Computa-tional Approaches to Discourse at the 61st Annual Meeting of the Association for Computational Linguistics (ACL CODI’23), July 9-14, 2023. Toronto, Canada.}

\bibitem[{Zhang et~al.(2023{\natexlab{b}})Zhang, Liu, and Zhang}]{zhang2023diffusum}
Haopeng Zhang, Xiao Liu, and Jiawei Zhang. 2023{\natexlab{b}}.
\newblock Diffusum: Generation enhanced extractive summarization with diffusion.
\newblock \emph{arXiv preprint arXiv:2305.01735}.

\bibitem[{Zhang et~al.(2023{\natexlab{c}})Zhang, Liu, and Zhang}]{zhang2023extractive}
Haopeng Zhang, Xiao Liu, and Jiawei Zhang. 2023{\natexlab{c}}.
\newblock Extractive summarization via chatgpt for faithful summary generation.
\newblock \emph{arXiv preprint arXiv:2304.04193}.

\bibitem[{Zhang et~al.(2023{\natexlab{d}})Zhang, Liu, and Zhang}]{zhang2023summit}
Haopeng Zhang, Xiao Liu, and Jiawei Zhang. 2023{\natexlab{d}}.
\newblock Summit: Iterative text summarization via chatgpt.
\newblock In \emph{Findings of the Association for Computational Linguistics: EMNLP 2023}, pages 10644--10657.

\bibitem[{Zhang et~al.(2024{\natexlab{a}})Zhang, Yu, and Zhang}]{zhang2024systematic}
Haopeng Zhang, Philip~S Yu, and Jiawei Zhang. 2024{\natexlab{a}}.
\newblock A systematic survey of text summarization: From statistical methods to large language models.
\newblock \emph{arXiv preprint arXiv:2406.11289}.

\bibitem[{Zhang et~al.(2020{\natexlab{a}})Zhang, Zhao, Saleh, and Liu}]{zhang2020pegasus}
Jingqing Zhang, Yao Zhao, Mohammad Saleh, and Peter Liu. 2020{\natexlab{a}}.
\newblock Pegasus: Pre-training with extracted gap-sentences for abstractive summarization.
\newblock In \emph{International conference on machine learning}, pages 11328--11339. PMLR.

\bibitem[{Zhang et~al.(2020{\natexlab{b}})Zhang, Kishore, Wu, Weinberger, and Artzi}]{zhang2020bertscoreevaluatingtextgeneration}
Tianyi Zhang, Varsha Kishore, Felix Wu, Kilian~Q. Weinberger, and Yoav Artzi. 2020{\natexlab{b}}.
\newblock \href {https://arxiv.org/abs/1904.09675} {Bertscore: Evaluating text generation with bert}.
\newblock \emph{Preprint}, arXiv:1904.09675.

\bibitem[{Zhang et~al.(2024{\natexlab{b}})Zhang, Ladhak, Durmus, Liang, McKeown, and Hashimoto}]{zhang2024benchmarking}
Tianyi Zhang, Faisal Ladhak, Esin Durmus, Percy Liang, Kathleen McKeown, and Tatsunori~B Hashimoto. 2024{\natexlab{b}}.
\newblock Benchmarking large language models for news summarization.
\newblock \emph{Transactions of the Association for Computational Linguistics}, 12:39--57.

\bibitem[{Zheng et~al.(2023)Zheng, Chiang, Sheng, Zhuang, Wu, Zhuang, Lin, Li, Li, Xing et~al.}]{zheng2023judging}
Lianmin Zheng, Wei-Lin Chiang, Ying Sheng, Siyuan Zhuang, Zhanghao Wu, Yonghao Zhuang, Zi~Lin, Zhuohan Li, Dacheng Li, Eric Xing, et~al. 2023.
\newblock Judging llm-as-a-judge with mt-bench and chatbot arena.
\newblock \emph{Advances in Neural Information Processing Systems}, 36:46595--46623.

\bibitem[{Zhong et~al.(2020)Zhong, Liu, Chen, Wang, Qiu, and Huang}]{zhong2020extractive}
Ming Zhong, Pengfei Liu, Yiran Chen, Danqing Wang, Xipeng Qiu, and Xuanjing Huang. 2020.
\newblock Extractive summarization as text matching.
\newblock \emph{arXiv preprint arXiv:2004.08795}.

\end{thebibliography}
